\documentclass[twoside]{article}

\usepackage[accepted]{aistats2020}
\usepackage{multicol}
\usepackage{microtype}
\usepackage{graphicx}
\usepackage{subfigure}
\usepackage{booktabs} 
\usepackage{wrapfig}
\usepackage{booktabs}
\usepackage[utf8]{inputenc} 
\usepackage[T1]{fontenc}    
\usepackage{url}            
\usepackage{amsfonts}       
\usepackage{nicefrac}       
\usepackage{subfigure}
\graphicspath{{figures/}}
\usepackage{xcolor}
\usepackage{adjustbox}
\usepackage{bm}
\usepackage{amsmath}
\usepackage{amssymb}
\usepackage{algorithm}
\usepackage{algorithmic}
\usepackage{multirow}
\usepackage{hyperref} 

\usepackage{tabularx}
\usepackage{ctable}
\usepackage{natbib}
\usepackage{indentfirst}
\newcommand{\RN}[1]{%
	\textup{\lowercase\expandafter{\it \romannumeral#1}}%
}

\newcommand{\ENC}{\text{Enc}}

\def\KL{\textsf{KL}} 

\ifx\definition\undefined
\newtheorem{definition}{Definition}
\fi

\ifx\remark\undefined
\newtheorem{remark}{Remark}
\fi

\usepackage{algorithm}
\usepackage{algorithmic}

\usepackage{amsmath}
\usepackage{amsfonts}
\usepackage{amssymb}

\newcommand{\beq}{\vspace{0mm}\begin{equation}}
\newcommand{\eeq}{\vspace{0mm}\end{equation}}
\newcommand{\beqs}{\vspace{0mm}\begin{eqnarray}}
\newcommand{\eeqs}{\vspace{0mm}\end{eqnarray}}
\newcommand{\barr}{\begin{array}}
\newcommand{\earr}{\end{array}}

\newcommand{\sv}{{\boldsymbol s}}

\newcommand{\uv}{{\boldsymbol u}}
\newcommand{\vv}{{\boldsymbol v}}

\newcommand{\Yv}{{\boldsymbol Y}}
\newcommand{\zv}{{\boldsymbol z}}

\usepackage[bottom]{footmisc}

\newcommand{\Amat}[0]{{{\bf A}}}

\newcommand{\Cmat}{{\bf C}}

\newcommand{\Qmat}[0]{{{\bf Q}}}

\newcommand{\Tmat}[0]{{{\bf T}}}

\newcommand{\nuv}{{\boldsymbol \nu}}
\newcommand{\muv}{{\boldsymbol \mu}}

\def\KL{\textsf{KL}} 

\def\argmin{\mathop{\rm arg\,min}}

\def\argmin{\mathop{\rm arg\,min}}

\def\KL{\mathop{{\rm KL}}}

\ifx\proof\undefined
\newenvironment{proof}{\par\noindent{\bf Proof\ }}{\hfill\BlackBox\\[2mm]}
\fi

\ifx\theorem\undefined
\newtheorem{theorem}{Theorem}
\fi
\ifx\example\undefined
\newtheorem{example}{Example}
\fi
\ifx\lemma\undefined
\newtheorem{lemma}[theorem]{Lemma}
\fi

\ifx\corollary\undefined
\newtheorem{corollary}[theorem]{Corollary}
\fi

\ifx\assumption\undefined
\newtheorem{assumption}{Assumption}
\fi

\ifx\definition\undefined
\newtheorem{definition}{Definition}
\fi

\ifx\proposition\undefined
\newtheorem{proposition}[theorem]{Proposition}
\fi

\ifx\remark\undefined
\newtheorem{remark}{Remark}
\fi

\ifx\conjecture\undefined
\newtheorem{conjecture}[theorem]{Conjecture}
\fi

\ifx\factoid\undefined
\newtheorem{factoid}[theorem]{Fact}
\fi

\ifx\axiom\undefined
\newtheorem{axiom}[theorem]{Axiom}
\fi

\begin{document}
	
	%
	
	%
	
	\twocolumn[
	
\aistatstitle{Nested-Wasserstein Self-Imitation Learning for Sequence Generation}


\aistatsauthor{  
		Ruiyi Zhang$^{1}$~~ Changyou Chen$^{2}$~~ Zhe Gan$^{3}$~~ Zheng Wen$^{4}$~~ Wenlin Wang$^{1}$~~ Lawrence Carin$^{1}$}
	
	\aistatsaddress{ $^{1}$Duke University~~ $^{2}$University at Buffalo~~ $^{3}$Microsoft Dynamics 365 AI Research~~ $^{4}$DeepMind \\
		\texttt{ryzhang@cs.duke.edu}  } ]

	\begin{abstract}
		Reinforcement learning (RL) has been widely studied for improving sequence-generation models. However, the conventional rewards used for RL training typically cannot capture sufficient semantic information and therefore render model bias. Further, the sparse and delayed rewards make RL exploration inefficient. To alleviate these issues, we propose the concept of nested-Wasserstein distance for distributional semantic matching.
		To further exploit it, a novel nested-Wasserstein self-imitation learning framework is developed, encouraging the model to exploit historical high-rewarded sequences for enhanced exploration and better semantic matching. Our solution can be understood as approximately executing proximal policy optimization with Wasserstein trust-regions. Experiments on a variety of unconditional and conditional sequence-generation tasks demonstrate the proposed approach consistently leads to improved performance.
		
	\end{abstract}
	\section{Introduction}
	Sequence generation is an important research topic in machine learning, covering a wide range of applications, including machine translation~\citep{bahdanau2014neural,rnnencdec,sutskever2014sequence}, image captioning \citep{anderson2017bottom, vinyals2015show, xu2015show}, and text summarization \citep{paulus2017deep, rush2015neural}. 
	Standard sequence generation follows an auto-regressive model design under maximum likelihood estimation (MLE) learning~\citep{huszar2015not,sutskever2014sequence,wiseman2016sequence}. That is, models are trained to maximize the expected log-likelihood of the next word conditioned on its preceding ground-truth partial sentence. However, when testing, the generated partial sequence is fed to the generator to draw the next token. Such a discrepancy between training and testing, commonly known as {\it exposure bias}, leads to accumulated approximation errors along the sequence-generation trajectory~\citep{bengio2015scheduled, ranzato2015sequence}. 
	
	To address exposure bias, reinforcement learning (RL) techniques have been introduced~\citep{ranzato2015sequence}. Unlike MLE, which only leverages training examples, RL can also exploit samples drawn from the current policy. Improvements are gained from reinforcing the training towards more-plausible generations, typically based on a user-specified reward function~\citep{ranzato2015sequence,yu2017seqgan}.
	However, the manually designed rewards often target specific desirable properties in sequence generation ({\it e.g.}, matching $n$-gram overlap between generated sequences and ground-truth references), which unintentionally induces extra bias and is often criticized as a bad proxy for human evaluation~\citep{wang2018no,hu2019makes}.
	Concerns have also been raised w.r.t. efficient exploration in sequence generation. In existing RL-based methods for sequence generation~\citep{bahdanau2016actor,ranzato2015sequence,rennie2016self}, all experiences are treated as equivalent. However, merely relying on policy samples to explore often leads to forgetting a high-reward trajectory, unless it can be re-sampled frequently~\citep{liang2018memory}. 
	This problem becomes severe in the sparse-reward setting in sequence generation, \textit{i.e.}, the reward is only available after the whole sentence is generated. 
	
	
	Motivated by the above observations, we present a novel nested-Wasserstein Self-Imitation Learning (WSIL) framework for sequence generation. Specifically,
	we propose the nested-Wasserstein distance, a generalization of the Wasserstein distance, and exploit it to measure distance between the behavior policy and the artificial policy defined by the replay buffer to encourage self-imitation. 
	The nested-Wasserstein distance is well suited for distributional semantic matching between two (sequence) distributions whose samples are still discrete distributions, as in the case of sequence generation.
	The proposed WSIL is inspired by and derived from the policy optimization with Wasserstein trust-regions~\citep{zhang2018policy}.
	It provides a novel reward function to match the generated sequences with the high-reward sequences in the replay buffer, encouraging distributional semantic matching rather than simple $n$-gram overlapping. 

	The main contributions of this paper are summarized as follows. ($i$) A novel nested-Wasserstein self-imitation learning framework is developed for sequence generation, \emph{exploiting} historical good explorations for better future \emph{exploration}. ($ii$) A novel nested-Wasserstein distance is introduced for sequence generation via distributional semantic matching, effectively alleviating the model training bias imposed by conventional rewards. ($iii$) Extensive empirical evaluation is performed on both unconditional and conditional text generation tasks, demonstrating consistent performance improvement over existing state-of-the-art approaches. 
	
	\vspace{-2mm}
	\section{Background}
	\paragraph{Sequence-generation model}
	\vspace{-2mm}
	We consider the problem of discrete sequence generation, which learns to generate a sequence $Y=(y_1, \ldots, y_T)\in\mathcal{Y}$ of length $T$, possibly conditioned on context $X$. Here each $y_t$ is a token from vocabulary $\mathcal{A}$. Pairs $(X,Y)$ are used for training a sequence-generation model. We are particularly interested in applications to text generation, where $Y$ is a sentence and each $y_t$ is a word. 
	%
	Starting from the initial state $\sv_0$,  
	a recurrent neural network (RNN) produces a sequence of states $(\sv_1,\ldots,\sv_T)$ given an input sequence-feature representation $(e(y_1), \dots, e(y_T))$, where $e(\cdot)$ denotes a word embedding mapping a token to its $d$-dimensional feature representation. 
	The states are recursively updated with a function known as the {\it cell}: $\sv_t=h_{\theta}(\sv_{t-1}, e(y_t))$, where $\theta$ denotes the model parameters. Popular implementations include Long Short-Term Memory (LSTM)~\citep{hochreiter1997long} and the Gated Recurrent Unit~(GRU)~\citep{rnnencdec}. 
	%
	%
	In order to generate sequence $Y^s$ from a (trained) model, one iteratively applies the following operations:
	\vspace{-5mm}
	\begin{align}
	{y}^s_{t+1} &\sim \text{Multi}(\text{softmax}(g(\sv_{t})))\,,\\\sv_t &= h(\sv_{t - 1}, e({y}^s_t))\,,
	\end{align}  
	\vspace{-8mm}
	
	where $\text{Multi}(\cdot)$ denotes a multinomial distribution. In conditional generation, 
	$\sv_0$ is initialized with $\ENC(X)$, where $\ENC(\cdot)$ encodes the relevant information from the context~\citep{bahdanau2016actor, rnnencdec}. For unconditional generation, one typically draws $\sv_0$ from a standard Gaussian distribution. 
	\vspace{-3mm}
	\paragraph{Sequence generation as an RL problem}
	Sequence generation can be considered as an RL problem with deterministic state transition and sparse reward. It 
	can be formulated as a Markov decision process (MDP) $\mathcal{M} = \langle\mathcal{S}, \mathcal{A}, P, r \rangle$, where 
	$\mathcal{S}$ is the state space, $\mathcal{A}$ is the action space, $P$ is the deterministic
	environment dynamics 
	and $r(\sv,y)$ is a reward function. 
	%
	%
	The policy $\pi_{\theta}$, parameterized by $\theta$, maps each state $\sv \in \mathcal{S}$ to a probability distribution over $\mathcal{A}$. The objective is to maximize the expected reward, defined as:
	\vspace{-3mm}
	\begin{equation}
	\begin{aligned}\label{eq:policy_learning}
	\hspace{-3mm} J(\pi_\theta) = \mathbb{E}_{Y\sim \pi_\theta}\left[r(Y)\right] = \sum_{t=1}^{T}\mathbb{E}_{(\sv_t, y_t)\sim \pi_\theta}\left[r(\sv_t, y_t)\right]\,,
	\end{aligned}
	\end{equation}
	
	\vspace{-5mm}
	where $Y \triangleq (\sv_1, y_1, \cdots, \sv_T, y_T)$ is a trajectory from policy $\pi_{\theta}$ with $y_t\in\mathcal{A}$, and $r(Y)$ represents the reward for a sentence $Y$, and $r(\sv_t, y_t)$ is the step-wise reward. 
	RL seeks to learn an optimal policy, that maximizes the expected total reward $J(\pi_\theta)$. 
	
	\vspace{-3mm}
	\paragraph{Optimal transport on discrete domains}
	The optimal transport (OT) distance $W_c(\muv, \nuv)$ is a discrepancy score that measures the distance between two probability distributions $\muv(\cdot)$ and $\nuv(\cdot)$ w.r.t. a cost function $c(\cdot,\cdot)$.
	Specifically, 
	we consider two discrete distributions  $\muv \triangleq \sum_{i=1}^n u_i \delta_{\zv_i}$ and $\nuv \triangleq \sum_{j=1}^m v_j \delta_{\zv'_j}$ with $\delta_{\zv}$ the Dirac delta function centered on $\zv$. 
	The weight vectors $\uv=\{u_i\}_{i=1}^n \in \Delta_n$ and $\vv=\{v_j\}_{j=1}^m \in \Delta_m$ respectively belong to the $n$ and $m$-dimensional simplex, \textit{i.e.}, $\sum_{i=1}^n u_i = \sum_{j=1}^m v_j = 1$.
	Accordingly, Wasserstein distance 
	is equivalent to solving the following minimization problem: 
	\vspace{-4mm}
	\begin{equation}
	\begin{aligned}
	\label{eq:wasserstein_distance}
	W_c(\muv,\nuv) &= \min_{\Tmat \in \Gamma(\muv,\nuv)}\sum^m_{i=1}\sum^n_{j=1}\Tmat_{ij} \cdot c(\zv_i,\zv'_j)\\
	&= \min_{\Tmat \in \Gamma(\muv,\nuv)} \,\, \langle \Tmat, \Cmat \rangle~,
	\end{aligned}
	\end{equation}
	
	\vspace{-5mm}
	where $\sum_{j=1}^n \Tmat_{ij}=\frac{1}{m}$ and $\sum_{i=1}^m \Tmat_{ij} = \frac{1}{n}$ are the constraints, $\langle \cdot, \cdot \rangle$ represents the Frobenius dot-product, and
	$\Cmat$ is the cost matrix defined by $\Cmat_{ij}=c(\zv_i,\zv'_j)$. Intuitively, the OT distance is the minimal cost of transporting mass from $\muv$ to $\nuv$. 
	\section{Distributional Semantic Matching}
	\label{sec: wasserstein_reward}
	We first consider evaluating the sentence from syntactic and semantic perspectives. Conventional metric rewards (\textit{e.g.}, BLEU) can capture the syntactic structure better, where the exact matching of words (or short phases) to the reference sequences is encouraged, which induces strong bias in many cases. As such, we focus on the semantic matching and propose the nested-Wasserstein distance, which defines
	the distance between two sequence distributions. Nested-Wasserstein distance provides a natural way to manifest semantic matching compared with the conventional rewards used in existing RL-based sequence models. Alternatively, we can train a discriminator to learn the reward model, but empirically it only rewards high-quality generations, even though they may be characterized by mode collapse~\citep{he2019lagging}. This undermines diversity, an important aspect in evaluation.
	
	\begin{figure}[t]
		\vspace{-2mm}
		\centering
		\includegraphics[width=0.95\linewidth]{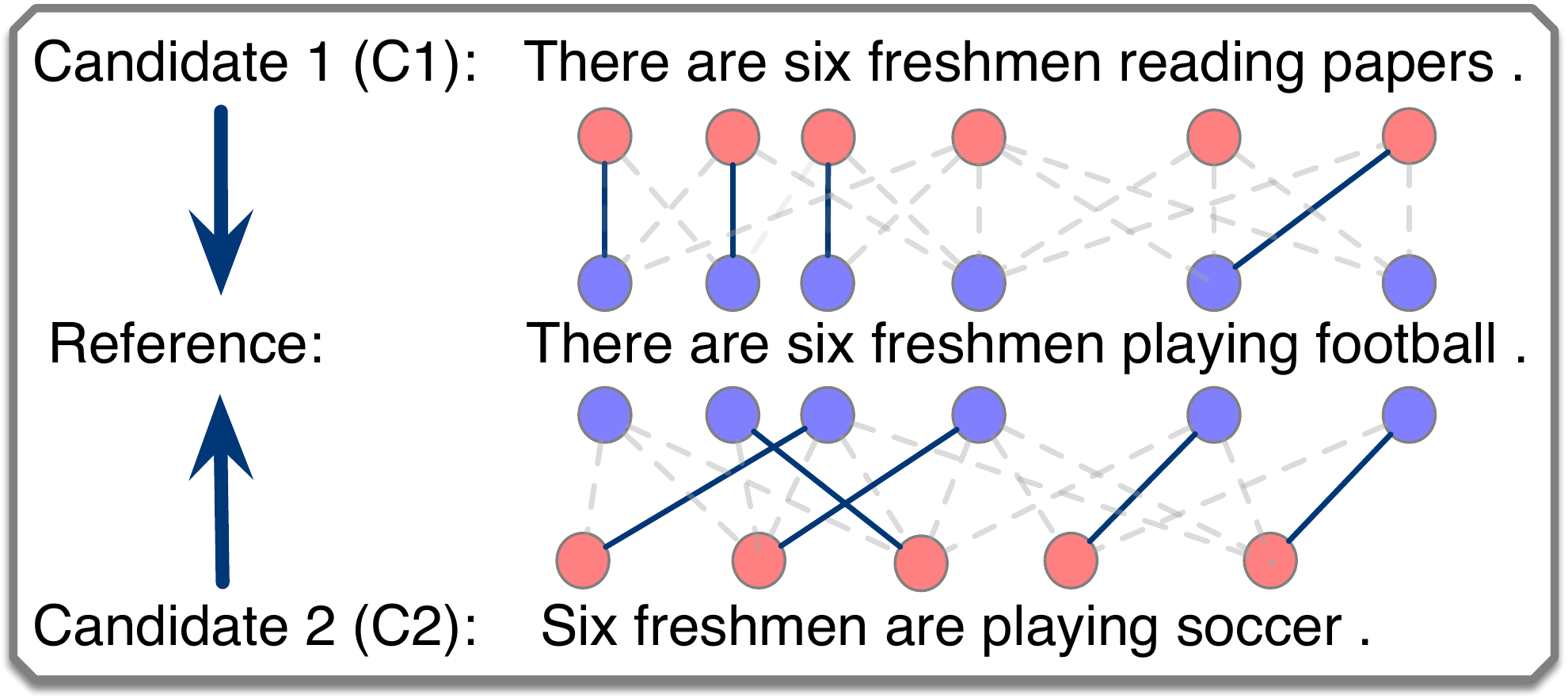}
		\vspace{-8mm}
	\end{figure}
	\begin{table}[t]
		\begin{center}
			\hspace{-0mm}
			\vspace{-2mm}
			\begin{adjustbox}{scale=0.89,tabular=lccccc,center}\\
				\toprule[1.2pt]
				& BLEU & ROUGE-L & CIDEr & Naive &Wasserstein\\
				\hline
				C1  &  {\bf 36.8}  & {\bf 50.0} & {\bf 163.7} &  \textbf{84.1} & 76.3 \\
				C2  & 0.0 & 35.8 & 55.9 & 42.5 & {\bf 80.1} \\
				\bottomrule[1.2pt]
			\end{adjustbox}
		\end{center}
		\caption{Comparison of different rewards in terms of the sequence-level (higher is better). The top figure illustrates the Wasserstein reward of comparing two candidate sentences with a reference sentence, which will automatically match semantically similar words. Dominant edges are shown in dark blue, determined by the optimal transport matrix $\Tmat$.}
		\label{table-exampe-metrics}
		\vspace{-2mm}
	\end{table}

	To better understand the issue, consider the example on sequence matching in Table \ref{table-exampe-metrics}. One can also use a naive way of semantic matching, \textit{i.e.,} measuring a distance between average word embeddings.
	It is clear that while the first candidate sentence has a similar syntactic structure to the reference, the second candidate sentence is more semantically consistent with the reference. However, popular hard-matching metrics~\citep{papineni2002bleu, vedantam2015cider} and the naive method  consistently indicate the first candidate is a better match to the reference. 
	The above contradiction can be alleviated if the reward metric is more semantic-aware. So motivated, the remainder of this section is devoted to a discussion of design and implementation of Wasserstein rewards. The general idea is to match the semantic features via minimizing the Wasserstein distance between hypothesis sentences and their references in the semantic space. 
	%
	A nested version of the Wasserstein distance arises when integrating the distributional semantic matching into the objective of sequence distribution matching.
	\vspace{-2mm}
	\begin{definition}\label{def:W}\textbf{(Wasserstein Distance between Sequence Pairs)}
		Consider sequence $Y=({y_1,\ldots,y_T})$ as a discrete distribution $p_{Y} = \frac{1}{T}\sum_{t} \delta_{e(y_t)}$ in the semantic space, with the length-normalized point mass placed at the word embedding, \textit{i.e.,} $\zv_t=e(y_t)$ of each token $y_t$ from the sequence $Y$.
		Given a hypothesis sequence $Y$ w.r.t. a reference sequence $Y'$, we define the Wasserstein distance as $W_c(p_{Y},p_{Y'})\triangleq \min_{\Tmat}\langle \Tmat, \Cmat \rangle$ between $p_{Y}$ and $p_{Y'}$ with cost $c(\zv,\zv')$. 
		When the {\it cosine distance} $c_{\cos}(\zv, \zv') = 1-\frac{\zv^\intercal\zv'}{\|\zv\|_2\|\zv'\|_2}$ is used as our cost, 
		we define the Wasserstein reward as $r_s(Y, Y^\prime)\triangleq\langle \Tmat^*, 1-\Cmat \rangle$, where $\Tmat^*$ is the optimal transport matrix.
	\end{definition}
	\begin{figure}[t]
		\vspace{-2mm}
		\centering
		\includegraphics[width=0.75\linewidth]{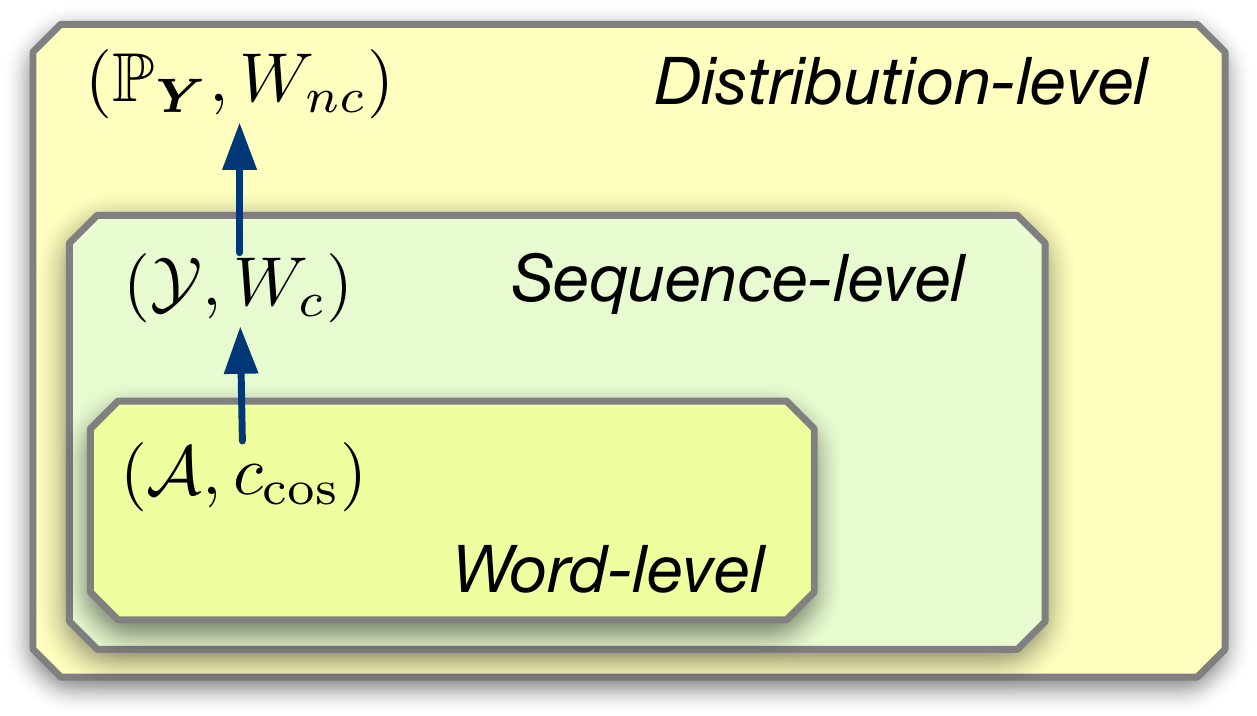}
		\vspace{-4mm}
		\caption{Illustration of nested-Wasserstein distance ($W_{nc}$) over distributions of sequences ($\mathbb{P}_{\Yv}$), showing how the distance is defined in a nested manner to measure distance of sequence distributions. $c_{\cos}$ is the word ground metric; $W_c$ is the sequence ground metric. }
		\label{fig: nested-Wasserstein}
		\vspace{-2mm}
	\end{figure}
	\vspace{-3mm}
	\paragraph{Nested-Wasserstein distance}
	Our ultimate goal is to measure distance between two policy distributions instead of sequence pairs. Given two sets of sequences from two policies, one aims to incorporate the semantic information between sequences into the distance measure. 
	To this end, we propose the nested-Wasserstein distance in Definition \ref{def:nW}. Figure \ref{fig: nested-Wasserstein} illustrates the nested-Wasserstein, considering both word- and sequence-level matching with Wasserstein distance.
	\vspace{-2mm}
	\begin{definition}[Nested-Wasserstein Distance]\label{def:nW}
		Consider two sets of sequences $\Yv = \{Y_i\}_{i=1}^K$ and $\Yv' = \{Y'_j\}_{j=1}^{K'}$ drawn from two sequence distributions $\mathbb{P}_{\Yv}$ and $\mathbb{P}_{\Yv'}$, where $K$ and $K'$ are the number of sequences in $\Yv$ and $\Yv'$. The nested-Wasserstein distance, denoted as $\mathcal{W}_{nc}(\mathbb{P}_{\Yv}, \mathbb{P}_{\Yv'})$, is a metric measuring the distance between $\mathbb{P}_{\Yv}$ and $\mathbb{P}_{\Yv'}$ defined in a nested manner:
		\vspace{-3mm}
		\begin{align}\label{eq:nW1}
		\mathcal{W}_{nc}(\mathbb{P}_{\Yv}, \mathbb{P}_{\Yv'})\triangleq \min_{T^s}\sum_{i=1}^K\sum_{j=1}^{K'} T^s_{ij} W_c(p_{Y_i},p_{Y_{j}'})~,
		\end{align}
		
		\vspace{-5mm}
		where $T_{ij}^\prime \geq 0$ satisfies $\sum_{i}T^s_{ij} = \frac{1}{K}$ and $\sum_{j}T^s_{ij} = \frac{1}{K'}$; and $W_c(\cdot, \cdot)$ denotes the c-Wasserstein distance defined in \eqref{eq:wasserstein_distance}.
		\vspace{-2mm}
	\end{definition}
	\vspace{-3mm}
	\begin{remark}
		The word ``nested'' comes from the definition in \eqref{eq:nW1}, which essentially consists of two nested levels of Wasserstein distances. The proposed nested-Wasserstein distance brings in the semantic information via the distance measure $W_c$ in the first level distance. Note that we have omitted the expectation over samples in \eqref{eq:nW1} for simplicity, as we essentially use a single set of samples to approximate $\mathcal{W}_{nc}(\cdot,\cdot)$ in algorithms. 
	\end{remark}
	\vspace{-4mm}
	\paragraph{Sample-based estimation of nested-Wasserstein distance}
	Computing the exact Wasserstein distance is computationally intractable~\citep{wgan, genevay2018learning, salimans2018improving}, let alone the proposed nested-Wasserstein distance. Fortunately, we can employ the recently proposed IPOT algorithm \citep{xie2018fast} to obtain an efficient approximation. Specifically, IPOT considers the following proximal gradient descent to solve the optimal transport matrix $\Tmat$ via iterative optimization, \textit{i.e.},
	$\Tmat^{(t+1)} = \argmin_{\Tmat\in\Pi(\muv,\nuv)} \left\{ \langle \Tmat, \Cmat\rangle + \gamma \cdot \mathbb{D}_{\KL}(\Tmat,\Tmat^{(t)}) \right\}$, 
	where $1/\gamma>0$ is the generalized step size and the generalized KL-divergence $\mathbb{D}_{\text{KL}}(\Tmat, \Tmat^{(t)})=\sum_{i,j} \Tmat_{ij}\log \frac{\Tmat_{ij}}{\Tmat^{(t)}_{ij}}-\sum_{i,j} \Tmat_{ij} + \sum_{i,j} \Tmat^{(t)}_{ij}$ is used as the proximity metric. Standard Sinkhorn iterations \citep{cuturi2013sinkhorn} are used to solve the above sub-problem. The IPOT was designed to approximately calculate the standard Wasserstein distance. Here we extend it to calculate the nested-Wasserstein distance by applying IPOT twice in a nested manner, {\it i.e.}, in the sequence and distribution levels, respectively. 
	The full approach of IPOT is summarized as Algorithm \ref{alg:ipot} in Appendix \ref{sec: impdetails}.
	\section{Nested-Wasserstein Self-Imitation Learning}
	\label{sec: nWSIL}
	Purely adopting the nested-Wasserstein distance as the reward in a standard policy-gradient method is not effective, because the syntactic information is missing. 
	Specifically,
	we consider sequences generated from a conditional behavior policy $\pi_{\theta,X}$, parameterized by $\theta$ with the conditional variable $X$. For example, in image captioning, each sequence is generated conditioned on a given image. For unconditional generation, the conditional variable is empty. 
	Instead of combining the rewards with different weights~\citep{liu2017improved, pasunuru2017reinforced}, we present the nested-Wasserstein Self-Imitation Learning (WSIL) framework, which provides a novel way of leveraging both syntactic (metric) and semantic (Wasserstein) information. 
	
	The overall idea of the proposed nested-Wasserstein self-imitation learning is to define a Wasserstein trust-region between the current policy (a.k.a.\! behavior policy) and the artificial policy defined by the replay buffer. Intuitively, the Wasserstein trust-region encourages the self-imitation of historical high-reward sequences, which provides semantic signals to guide training, in addition to the stabilizing effect from trust-region optimization. 
 Furthermore, a replay buffer is used to store high-reward historical sequences, whose induced conditional policy is denoted  $\pi_{\mathcal{B}, X}$. 
	Our new objective function with a Wasserstein trust-region is defined as:
	\vspace{-1mm}
	\begin{equation}
	\begin{aligned}\label{eq:theorry_self_learning}
	J(\pi_\theta) =\mathbb{E}_{X\sim p_d}&
	\{\mathbb{E}_{Y^s\sim\pi_{\theta,X}}\left[r(Y^s) \right] \\&- \lambda \cdot \mathcal{W}_{nc}(\pi_{\theta,X}, \pi_{\mathcal{B},X})\}~,
	\end{aligned}
	\end{equation}
	
	\vspace{-2mm}
	where $\mathcal{W}_{nc}$ is the nested-Wasserstein distance defined in Definition \ref{def:nW}, and $r(\cdot)$ can be a metric reward between $Y^s$ and the ground-truth references $\Yv$. With a little abuse of notation, but for conciseness, we use $\pi_\theta$ to denote both the policy and the distribution over the sequences. Distinct from classic trust-region policy optimization, which defines the trust region based on KL-divergence~\citep{schulman2015trust}, WSIL defines the trust region based on the nested-Wasserstein distance between the behavior policy $\pi_{\theta,X}$ and the artificial policy $\pi_{\mathcal{B},X}$.
	Note when $K=K'=1$, the nested Wasserstein distance degenerates to the definition of Wasserstein distance between two sequences.

\begin{figure}[t!]
		\hspace{0mm}\centerline{\includegraphics[width=\linewidth]{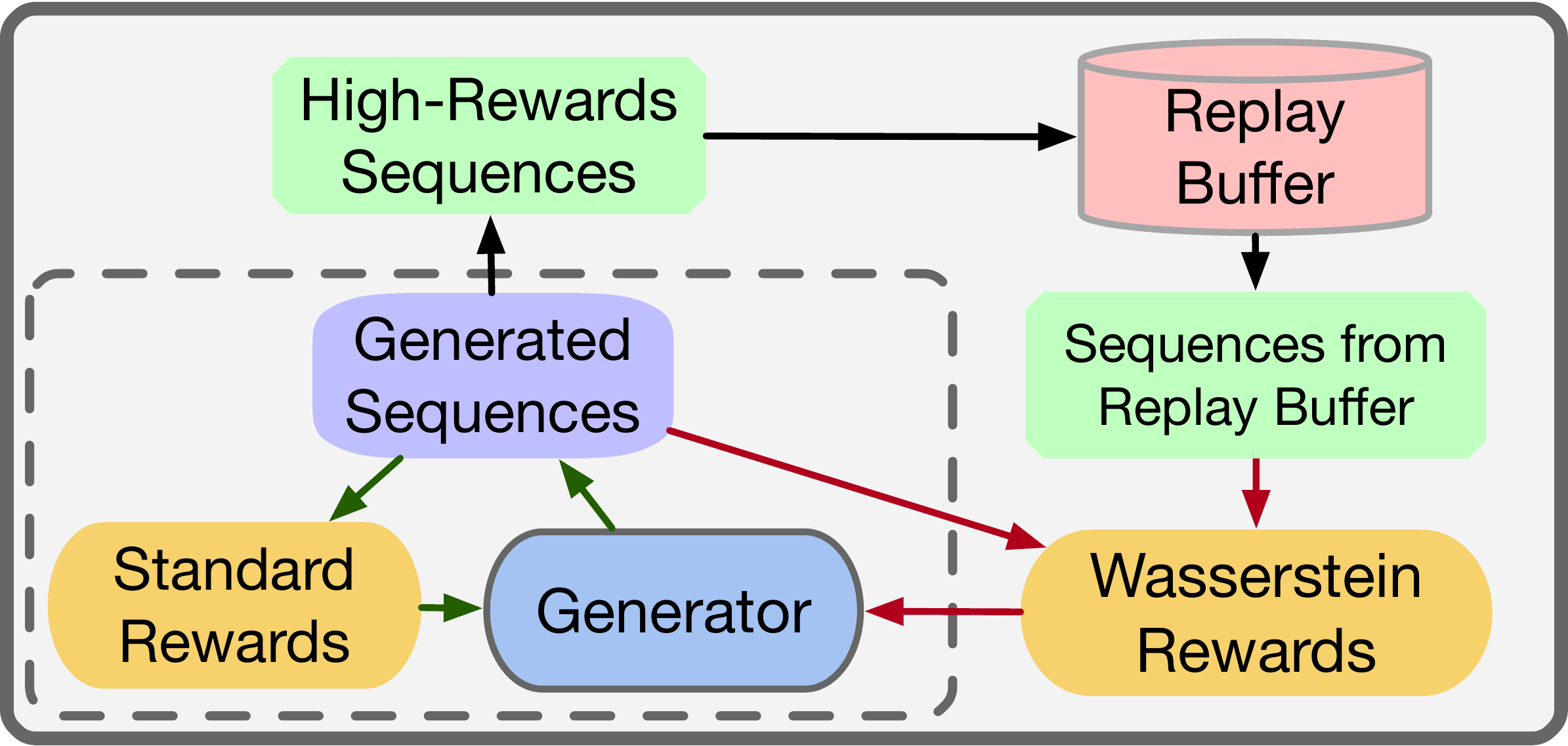}}
		 \vspace{-6mm}
		\caption{Illustration of the proposed nested-Wasserstein Self-Imitation Learning (WSIL) framework, where Wasserstein self-imitation rewards are defined to encourage the generator to imitate samples from the replay buffer. The standard RL framework is given in the gray dotted box.}
		\label{fig:framework}
		\vspace{-3mm}
	\end{figure}	
	
	\vspace{-2mm}
	\begin{remark}{\textbf{Unconditional Generation:}}
		By considering samples (features) themselves as discrete distributions, we replace the mean square difference over features of sequence pairs, \textit{i.e.}, Euclidean norm, with the Wasserstein distance. Then for the distributions of sequences, we again adopt the Wasserstein distance as in WGAN~\citep{wgan} but in the discrete domain. Thus, the Wasserstein distance is defined in a nested manner. 
	\end{remark}
	\vspace{-3mm}
	\begin{remark}{\textbf{Conditional Generation:}}
		 We replace the exact matching of sequence pairs with metric rewards in RL training, with the Wasserstein distance. In this case, we are matching two conditional distributions with Wasserstein distance, instead of matching the generated sentence with all reference sentences by average. This is a more suitable way as a generated sentence does not necessarily need to match all the references.
	\end{remark}
	\vspace{-2mm}
For simplicity, we sometimes omit the first expectation $\mathbb{E}_{X\sim p_d}$. With the proposed nested Wasserstein distance, we propose the Wasserstein self-imitation scheme in \eqref{eq:theorry_self_learning}, as illustrated in Figure~\ref{fig:framework}.
%
	We seek to use historical high-reward sequences to define a ``\emph{self-imitation}'' reward function, which is then combined with the original reward function to update the generator with policy gradient methods. Intuitively, higher self-imitation rewards are achieved when the generated sequences are close to historical high-reward sequences. Thus the generator is guided to perform self imitation and we call this method indirect nested-Wasserstein self-imitation learning (WSIL-I). The word ``\emph{indirect}'' comes from the mechanism that historical sequences interact with the policy indirectly via the {\em self-imitation} reward. 
	%
	
	WSIL-I incorporates a self-imitation reward, denoted as $r_{s}(Y^s, Y^b)$, into the objective function. Here $Y^b$ denotes a sample from the replay buffer and $Y^{s}$ denotes a sample from the current policy. To this end, we replace the Wasserstein distance $W_c$ in the nested-Wasserstein distance with $r_{s}(Y^s, Y^b)$ in the general objective \eqref{eq:theorry_self_learning}. 
	Specifically, we define the two sets of sample sequences from $\pi_{\theta,X}$ and $\pi_{\mathcal{B},X}$ to be $\Yv^s \triangleq\{Y^s_i\}_{i=1}^K$ and $\Yv^b \triangleq \{Y_j^b\}_{j=1}^{K'}$, with sizes of $K$ and $K^\prime$, respectively. Here $Y_i^s \sim \pi_{\theta,X}$ and $Y_j^b \sim \pi_{\mathcal{B},X}$, $\forall j$. $\{Y^s_i\}_{i=1}^K$ and $\Yv^b$ will be used in calculating the nested-Wasserstein distance. Let $r_{ns}(Y^s_i, \Yv^b)\triangleq \sum_j T^s_{ij} r_{s}(Y^s_i, Y^b_j)$ be the nested-Wasserstein reward, with $\Tmat^s = \{T^s_{ij}\}$ the optimal weights in distribution-level.
	Based on \eqref{eq:theorry_self_learning}, the objective of WSIL-I is adapted to be:
	\vspace{-1mm}
	\begin{align}
	\hspace{-4mm}
	J_I(\pi_\theta) &\triangleq \mathbb{E}_{X\sim p_d}\mathbb{E}_{Y^s\sim\pi_{\theta,X}}\left[r(Y^s) + \lambda r_{ns}(Y^s, \Yv^b)\right]\,,
	\label{eq:prop1}
	\end{align}
	
	\vspace{-2mm}
	where $r$ is the original RL reward; $r_{ns}$ is the nested-Wasserstein reward.
	Since not all historically explored samples are helpful for updating the current policy, we only consider a subset of the high-reward sequences when performing self-imitation. Using $K$ trajectories sampled \textit{i.i.d.} from $\pi_\theta$ and introducing a baseline $b$, the gradient estimate of WSIL-I is expressed as:
	\vspace{-3mm}
	\begin{equation}
	\begin{aligned}
	\nabla_{\theta}J_I(\pi_\theta)\approx&-\sum_{k=1}^K [(r(Y^s_k)-b) \nabla_{\theta}\log \pi_{\theta}(Y^s_k)\\
	&+\lambda
	r_{ns}(Y^s_k, \Yv^b)\nabla_{\theta} \log \pi_{\theta}(Y_k^s)]
	\,.
	\label{eq:iwsil}
	\end{aligned}
	\end{equation}
	
	\vspace{-3mm}
	
	In practice, $\mathcal{I}\left[r(Y^b)>r(Y^s)\right]$ will be combined with the nested-Wasserstein rewards, where 
	$\mathcal{I}(\cdot)=1$ if the condition is satisfied, and 0 otherwise; 
	$b$ is the baseline to stabilize training. If the reward of a historical high-reward sequence is greater than the current one (\emph{i.e.}, $r(Y^b)> r(Y^s)$), the generator learns to imitate this high-reward sequence. Otherwise, the update based on the historical sequence is not performed due to the $\mathcal{I}(\cdot)$ operator. This encourages the agent to only imitate its good historical explorations. We have also developed another way to implement (direct) WSIL (WSIL-D) as discussed in the Appendix A. Algorithm \ref{algo:self_algo} describes the general implementation procedure of the WSIL. 
	
	\begin{algorithm}[htp]
	\caption{Nested-Wasserstein Self-Imitation.}
	\label{algo:self_algo}
	\begin{algorithmic}
		\REQUIRE Generator policy $\pi_{\theta}$; a sequence dataset $\mathcal{D}=\{Y_{1 \ldots T}\}_{1}^{N}$; a possibly empty condition $\mathcal{X}=\{X\}_{1}^{N}$.
		\STATE Initialize $\pi_{\theta}$ and replay buffer $\mathcal{B}$.
		\STATE Pretrain generator $\pi_{\theta}$ with MLE.
		\REPEAT 
		\STATE{Generate $K$ sequences $\Yv^s=\{Y^s_k\}_{k=1}^{K}$, where $Y_k^s\sim \pi_{\theta}$.}
		\STATE{Update replay buffer $\mathcal{B}$ using $\Yv^s$.}
		\IF{Self-Imitation}
		\STATE{Sample $K'$ sequences $\Yv^b=\{Y^b_j\}_{j=1}^{K'}$, where $Y_j^b \sim \pi_\mathcal{B}$.}
		\STATE{Estimate the OT matrix $\Tmat$ and $\Tmat^s$ via IPOT}
		\STATE{Compute $r_{ns}(Y^s_k, \Yv^b)$ and update $\pi_{\theta}$ with \eqref{eq:iwsil}.}
		\ELSE
		\STATE{Update the generator $\pi_{\theta}$ with \eqref{eq:policy_learning} using $\Yv^s$.}
		\ENDIF
		\UNTIL{Algorithm converges}
	\end{algorithmic}
\end{algorithm}	
	
	\vspace{-2mm}
	\paragraph{Exploration Efficiency}  
	The exploration space of MLE is the examples in the training set~\citep{tan2018connecting}, \textit{i.e.,} no exploration is performed in supervised training. In contrast, standard policy optimization~\citep{ranzato2015sequence} basically allows the whole exploration space. However, the exploration may become inefficient since it may be too flexible, and 
	some good sequences observed in history tend to be less explored and imitated due to the sparse rewards. Our proposed WSIL aims to provide more efficient and systematic exploration. It allows the whole-space exploration, but re-weights the exploration space to focus more on the exploration that may provide better performance with the Wasserstein trust-region.
	\vspace{-2mm}
	\begin{figure}[htp]
		\vspace{-2mm}
		\centering
		\includegraphics[width=0.9\linewidth]{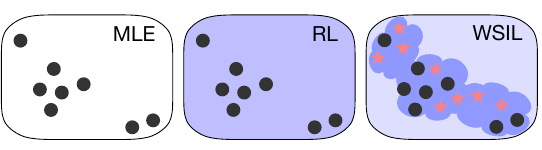}
		\vspace{-3mm}
		\caption{\small Exploration space of different methods. Circle: ground truth; Star: high-reward sequences.}
		\label{fig:toygan}
		\vspace{-3mm}
	\end{figure}

	\vspace{-1mm}
	\paragraph{Increasing Self-Imitation} According to the theory of Wasserstein gradient flows~\citep{Villani:08},  $1/\lambda$ can be interpreted as a generalized decaying learning rate. With more explorations, $\lambda$ becomes larger, and the algorithm should focus more on the self-imitation learning, providing a guideline to balance the standard RL training and self-imitation learning. More details are provided in Appendix \ref{sec: impdetails}. Practically, nested-Wasserstein provides weak supervision focusing on semantic matching, which is reasonable since the historical high-reward sequences contain some noises. 
	
	\vspace{-2mm}
	\section{Related Work}
	\vspace{-2mm}
	\paragraph{Optimal transport} \citet{kusner2015word} proposed the \textit{word mover's distance} (WMD) and first applied optimal transport (OT) to NLP; OT has also been employed to improve topic modeling~\citep{huang2016supervised}.
	The transportation cost is usually defined as Euclidean distance, and OT distance is approximated by solving a Kantorovich-Rubinstein dual~\citep{gulrajani2017improved} or a less-accurate lower bound~\citep{kusner2015word}.
	 \citet{yurochkin2019hierarchical} proposed a hierarchical OT representation for document, but the hierarchy was in word- and topic-level based on the WMD. Our work considers nested-Wasserstein distance, presenting an efficient IPOT-based implementation for OT distance approximation~\citep{xie2018fast}, successfully using it to guide sequence generation.
	\vspace{-3mm}
	\paragraph{Self-Imitation Learning} Experience replay has been widely considered in RL. Deterministic policy gradient~\citep{silver2014deterministic, lillicrap2015continuous} performs experience replay, but is limited to continuous control. 
	Actor-critic approaches \citep{konda2000actor} can also utilize a replay buffer to improve performance. 
	Prioritized experience replay~\citep{schaul2015prioritized} samples trajectories based on the time-difference error, and we adopt it in our implementation. These approaches indiscriminately buffer {\em all} experiences, while the approach proposed here only buffers high-reward experience. Further, episodic control~\citep{lengyel2008hippocampal} can be regarded as an extreme way of exploiting past experience, trying to reproduce its best past decisions, but retrieving states leads to poor efficiency and generalization in testing. Self-imitation learning was first applied in Atari games and Mujoco \citep{oh2018self,gangwani2018learning}, reporting performance improvement w.r.t. sparse rewards. Compared with that work, our solution considers a novel self-imitation learning scheme in the context of sequence generation. 
	\vspace{-8mm}
	\paragraph{RL for Sequence Generation} RL techniques have been explored in detail for sequence generation. For example, a Seq2Seq model can be trained by directly optimizing  BLEU/ROUGE scores via policy gradient~\citep{ranzato2015sequence,bahdanau2016actor}. Furthermore, \citet{rennie2016self} baselines the actor with the reward of a greedy-decoding sequence for the REINFORCE method. Model-based RL and hierarchical RL have also been studied for sequence generation~\citep{zhang2018sequence,huang2019hierarchically}. 
	Further, a learned discriminator (or, critic) can also be used to provide sequence-level guidance. By constructing different objectives, previous work~\citep{yu2017seqgan,lin2017adversarial, guo2017long, fedus2018maskgan} combines the policy-gradient algorithm with the original GAN training procedure. However, mode-collapse problems make the training of these methods challenging. 
	By contrast, we propose the use of self-imitation learning, and maintain a replay buffer to exploit past good explorations. 
	\begin{table*}[h]
	\vspace{-4mm}
	\centering
	\begin{adjustbox}{scale=0.91, tabular=l rccc  rccc rcc,center}
		\\ \toprule[1.2pt]
		\textbf{Method} & \textbf{Test-BLEU-2} & \textbf{3} & \textbf{4} & \textbf{5} & \textbf{Self-BLEU-2} & \textbf{3} & \textbf{4} \\
		\midrule
		MLE~~\citep{caccia2018language}   & 0.902 & 0.706 & 0.470 & 0.392 & 0.787 &0.646 & 0.485 \\
		SeqGAN \citep{yu2017seqgan}  & 0.820 & 0.604 & 0.361 & 0.211 & 0.807 & 0.577 & 0.278 \\
		RankGAN \citep{lin2017adversarial} & 0.852 & 0.637 & 0.389 & 0.248 & 0.822 & 0.592 & 0.230 \\
		TextGAN \citep{zhang2017adversarial} & 0.910 & 0.728 & 0.484 & 0.306 & 0.806 & 0.548 & 0.217  \\
		FMGAN \citep{chen2018adversarial} & 0.911 & 0.782 & 0.584 & 0.382 & 0.834 & 0.643 & 0.405  \\
		LeakGAN \citep{guo2017long} & 0.922 & 0.797 & 0.602 & 0.416 & 0.912 & 0.825 & 0.689  \\
		\hline
		WSIL-D (ours) & 0.917 & 0.774 & 0.576 & 0.393 & 0.797 & 0.569 & 0.284\\
		WSIL-I (ours) & 0.922 & 0.778 & 0.576 & 0.396 & 0.813 & 0.600 & 0.326 \\
		\bottomrule[1.2pt]
	\end{adjustbox}
	\vspace{-4mm}
	\caption{Test-BLEU ($\uparrow$) and Self-BLEU ($\downarrow$) scores on Image COCO.}
	\label{tab:coco}
	\vspace{-4mm}
\end{table*}

\begin{table*}[h]
	\vspace{-2mm}
	\centering
	\begin{adjustbox}{scale=0.91,tabular=l rccc  rcc rcc,center}
		\\ \toprule[1.2pt]
		\textbf{Method} & \textbf{Test-BLEU-2} & \textbf{3} & \textbf{4} & \textbf{5} & \textbf{Self-BLEU-2} & \textbf{3} & \textbf{4}  \\
		\midrule
		MLE~\citep{caccia2018language} & 0.905 & 0.701 & 0.464 & 0.278 & 0.764  & 0.522 & 0.295 \\
		SeqGAN \citep{yu2017seqgan}  & 0.630 & 0.354 & 0.164 & 0.087 & 0.728 & 0.411 & 0.139  \\
		RankGAN \citep{lin2017adversarial} & 0.723 & 0.440 & 0.210 & 0.107 & 0.672 & 0.346 & 0.119 \\
		TextGAN \citep{zhang2017adversarial} & 0.777 & 0.529 & 0.305 & 0.161 & 0.806 & 0.662 & 0.448 \\
		FMGAN \citep{chen2018adversarial} & 0.913 & 0.751 & 0.512 & 0.315 & 0.830 & 0.682 & 0.427  \\
		LeakGAN \citep{guo2017long} & 0.923 & 0.757 & 0.546 & 0.335 & 0.837 & 0.683 & 0.513 \\
		\hline
		SIL-D (ours) & 0.875 & 0.634 & 0.401 & 0.243 & 0.724 & 0.466 & 0.256 \\
		SIL-I (ours) & 0.869 & 0.633 & 0.399 & 0.242 & 0.710 & 0.455 & 0.263 \\
		WSIL-D (ours) & 0.931 & 0.736 & 0.503 & 0.317 & 0.795 & 0.553 & 0.299 \\
		WSIL-I (ours) & 0.926 & 0.726 & 0.492 & 0.307 & 0.815 & 0.595 & 0.380 \\
		\bottomrule[1.2pt]
	\end{adjustbox}
	\vspace{-4mm}
	\caption{Test-BLEU ($\uparrow$) and Self-BLEU ($\downarrow$) scores on EMNLP2017 WMT News.}
	\label{tab:wmt}
	\vspace{-4mm}
\end{table*}
\section{Experiments}
%
We evaluate the proposed method on both unconditional and conditional text-generation tasks, considering standard benchmark datasets. 
Our approach achieves state-of-the-art results on unconditional text generation and video captioning. We also observed improved performance on image captioning though relying on much simpler features compared to prior state-of-the-art methods. 
We also perform ablation studies to understand the improvements brought by self-imitation and Wasserstein rewards individually. 
Details of the datasets, experimental setup and
model architectures are provided in Appendix C. 
\vspace{-3mm}
\paragraph{Implementation Details} A few key techniques are required for successful model training. (\RN{1}) The reward from a greedy-decoding sentence is used as the baseline~\citep{rennie2016self} in conditional text generation; in unconditional text generation, a constant baseline is used.  
(\RN{2}) A single large replay buffer is maintained for unconditional generation, and multiple replay buffers are maintained for different conditions in conditional generation. (\RN{3}) For each pair of sentences, the shorter one should be padded to the same length as the longer one for a balanced optimal transport, which is a key implementation technique. 

\vspace{-3mm}
\paragraph{Demonstration of nested-Wasserstein}

\begin{figure}[t!]
	\centering
	\includegraphics[width=\linewidth]{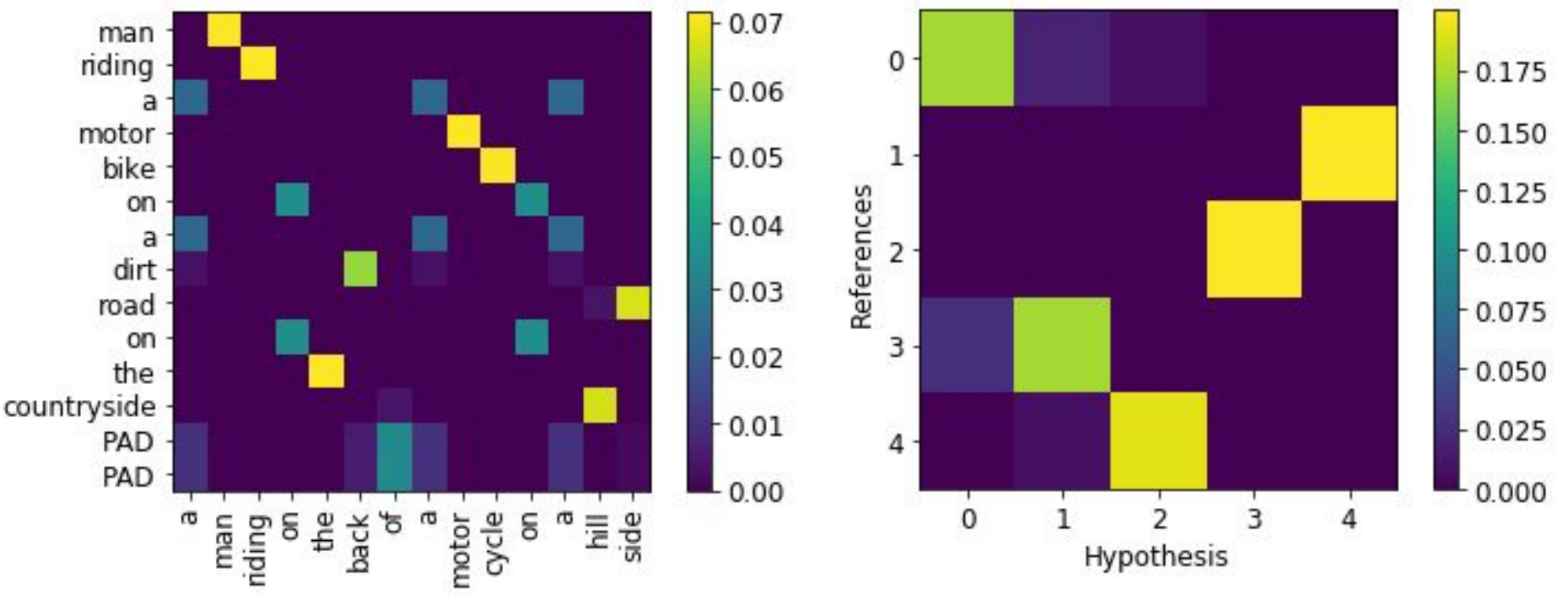}
	\vspace{-8.5mm}
	\caption{{Demonstration of nested-Wasserstein distance in word-level (left) and sentence-level (right).}}
	\label{fig:demo}
	\vspace{-3mm}
\end{figure}

\begin{figure}[t!]
	\centering
	\includegraphics[width=\linewidth]{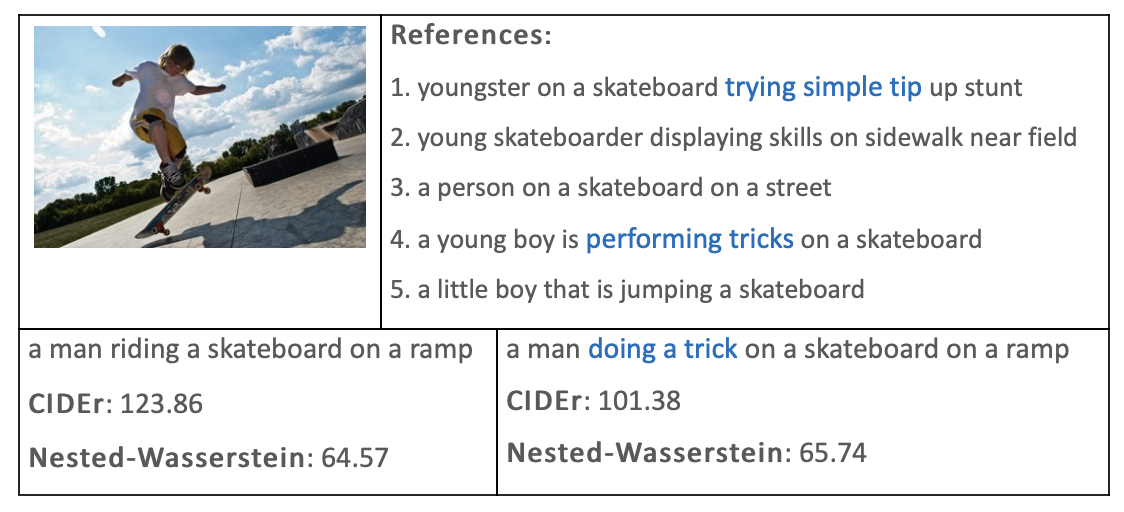}
	\vspace{-8.5mm}
	\caption{{An example of image captioning. The right generated sentence is better but given a lower CIDEr.}}
	\label{fig:demo2}
\end{figure}

Figure \ref{fig:demo} shows the optimal matching in word-level ($\Tmat$) and sentence-level ($\Tmat^s$). It is interesting to see that all similar words (\textit{e.g.}, bike and cycle) are matched with each other (higher weights), which cannot be achieved via exact hard-matching metrics. At the distribution-level, we show an example in captioning tasks, where we have five reference and hypothesis sentences. Traditional methods will match a hypothesis sentence to each of the references and average over them; while our method performs \textit{distributional semantic matching}, \textit{i.e.,} only matching similar references instead of all of them. For example, the third hypothesis is almost matched with the fifth reference, because they are more similar. This is reasonable, because the references are usually very different, and equivalently matching with all of them is confusing for the generator. As shown in Figure \ref{fig:demo2}, CIDEr focuses more on the locality fluency and equivalent matching with all references, while nested-Wasserstein performs distributional semantic matching. More examples are provided in the Appendix.   
\vspace{-2mm}
\vspace{-2mm}
\subsection{Unconditional Text Generation} 
\vspace{-1mm}
    We compare our approach with a number of related RL-based GAN models for unconditional text generation~\citep{guo2017long,lin2017adversarial,yu2017seqgan,zhang2017adversarial}. Our implementation is developed based on the LeakGAN model, by incorporating Wasserstein self-imitation learning. All baseline experiments are performed on the texygen platform
    ~\citep{zhu2018texygen}.
	The corpus-level BLEU score is employed to  evaluate the generated sentences. Specifically, we follow the strategy in \citet{yu2017seqgan,guo2017long} and adopt the BLEU score, referenced by test set (test-BLEU) and themselves (self-BLEU) to evaluate the quality of generated samples. 
    Test-BLEU evaluates the goodness of generated samples, and self-BLEU measures their diversity.
	The BLEU scores for 1000 generated sentences are averaged to obtain the final score for each model. A good generator should achieve both a high test-BLEU score and a low self-BLEU score. 
	Following previous work~\citep{guo2017long}, we test the proposed method on the short and long text generation on Image COCO and EMNLP2017 WMT News datasets. The BLEU scores with different methods are provided in Tables~\ref{tab:coco} and \ref{tab:wmt}. 
\vspace{-3mm}
\paragraph{Analysis} Compared with other methods, LeakGAN, WSIL-D and WSIL-I achieve comparable test-BLEU scores, demonstrating high-quality generated sentences. However, LeakGAN tends to over-fit on training data, leading to much higher (worse) self-BLEU scores. Our proposed methods, by contrast, show good diversity of the generated text with lower self-BLEU scores. 
Other baselines obtain both low self-BLEU and test-BLEU scores, leading to more random generations.
\begin{table*}[htp]
	\centering
	\begin{minipage}{.5\textwidth}
		\vspace{-2.5mm}
		\begin{center}
			\begin{adjustbox}{scale=0.63,tabular=lcccccc,center}
				\\ \toprule[1.2pt]
				\textbf{Method} & \textbf{BLEU-4} & \textbf{METEOR} & \textbf{ROUGE-L} & \textbf{CIDEr} \\ 
				\hline
				ED-LG~\citep{yao2015describing} & 35.2 & 25.2 & - & -  \\
				SA-LSTM~\citep{xu2016msr} & 36.6 & 25.9 & - & - \\
				SCST~\citep{pasunuru2017reinforced} & {40.5} & {28.4} & {61.4}  & {51.7}\\
				MBP~\citep{wang2018video} & 41.3 & 28.7 & 61.7 & 48.0 \\
				\hline
				\multicolumn{2}{c}{\textsc{Our Implementations}}\\
				\hline
				MLE & 39.2 & 27.8 & 59.8 & 46.6 \\
				MIXER~\citep{ranzato2015sequence} & 40.2 & 27.9 & 60.8 & 50.3\\
				SCST~\citep{rennie2016self} & 40.7 & 27.9 & 61.6 & 51.3 \\
				WSIL-D & \textbf{42.5} & \textbf{29.0} & \textbf{62.4} & 52.1\\
				WSIL-I & 41.6 & 28.4 & 62.0 & \textbf{52.2} \\
				\bottomrule[1.2pt]
			\end{adjustbox}
			\vspace{-4mm}
			\caption{Video captioning results on MSR-VTT.}
			\label{tab:videocaptioning}
		\end{center}
	\end{minipage}%
	\begin{minipage}{0.5\textwidth}
		\begin{center}
			\vspace{-2.5mm}
			\begin{adjustbox}{scale=0.63,tabular=lcccccccc,center}
				\\ \toprule[1.2pt]
				\textbf{Method} & \textbf{BLEU-4} & \textbf{METEOR} & \textbf{ROUGE-L} & \textbf{CIDEr}\\
				\hline
				S \& T~\citep{vinyals2015show}  & 27.7 & 23.7 & - &85.5\\
				OT~\citep{chen2019improving} &  31.0 & 24.6 & - & 94.7\\
				Adaptive~\citep{lu2017knowing} &  33.2 & 26.6 & - & 108.5\\
				TD~\citep{anderson2017bottom} & 33.3 & 26.3 & 55.3 & 111.4\\
				\hline
				\multicolumn{3}{c}{\textsc{Our Implementations}} \\
				\hline	
				MLE  & 28.8 & 24.4 & 52.0 & 91.3\\
				MIXER~\citep{ranzato2015sequence} & 30.8 & 24.7 & 52.9 & 101.2\\
				SCST~\citep{rennie2016self} & \textbf{32.1} & 25.4 & 53.9 & 105.5\\
				WSIL-D & 31.8 & \textbf{25.7} & \textbf{54.0} & 107.4\\
				WSIL-I & 32.0 & 25.6 & 53.9 & \textbf{107.6} \\	
				\bottomrule[1.2pt]
			\end{adjustbox}
			\vspace{-4mm}
			\caption{Image captioning results on COCO.}
			\label{tab:captioning}
		\end{center}
	\end{minipage}
	\vspace{-5mm}    
\end{table*}
\vspace{-3mm}
\paragraph{Ablation Study}
We conduct ablation studies on EMNLP2017 WMT news to investigate the improvements brought by each part of WSIL. First, we test the benefits of using two types of self-imitation schemes. We compare RL training with ($\RN{1})$ self-imitation (SIL-D and SIL-I), where only a replay buffer and conventional matching (features extracted from a neural network) are employed; and ($\RN{2})$ Wasserstein self-imitation (WSIL-D and WSIL-I). Results are shown in Table~\ref{tab:wmt}. We observe that the self-imitation strategy, with specific replay buffer construction, can alleviate the discrepancies between reward model bias and conventional rewards (\textit{e.g.}, self-BLEU). Without Wasserstein rewards, we achieve lower self-BLEU at the sacrifice of test-BLEU. When combining with Wasserstein rewards, WSIL-D and WSIL-I show superior performance relative to the baselines. The random generated samples in Appendix \ref{sec: uncon_examples} and human evaluations further validate this.

\begin{figure}[t] \centering
	\begin{tabular}{cc}
		\hspace{-3mm}
		\includegraphics[width=0.5\linewidth]{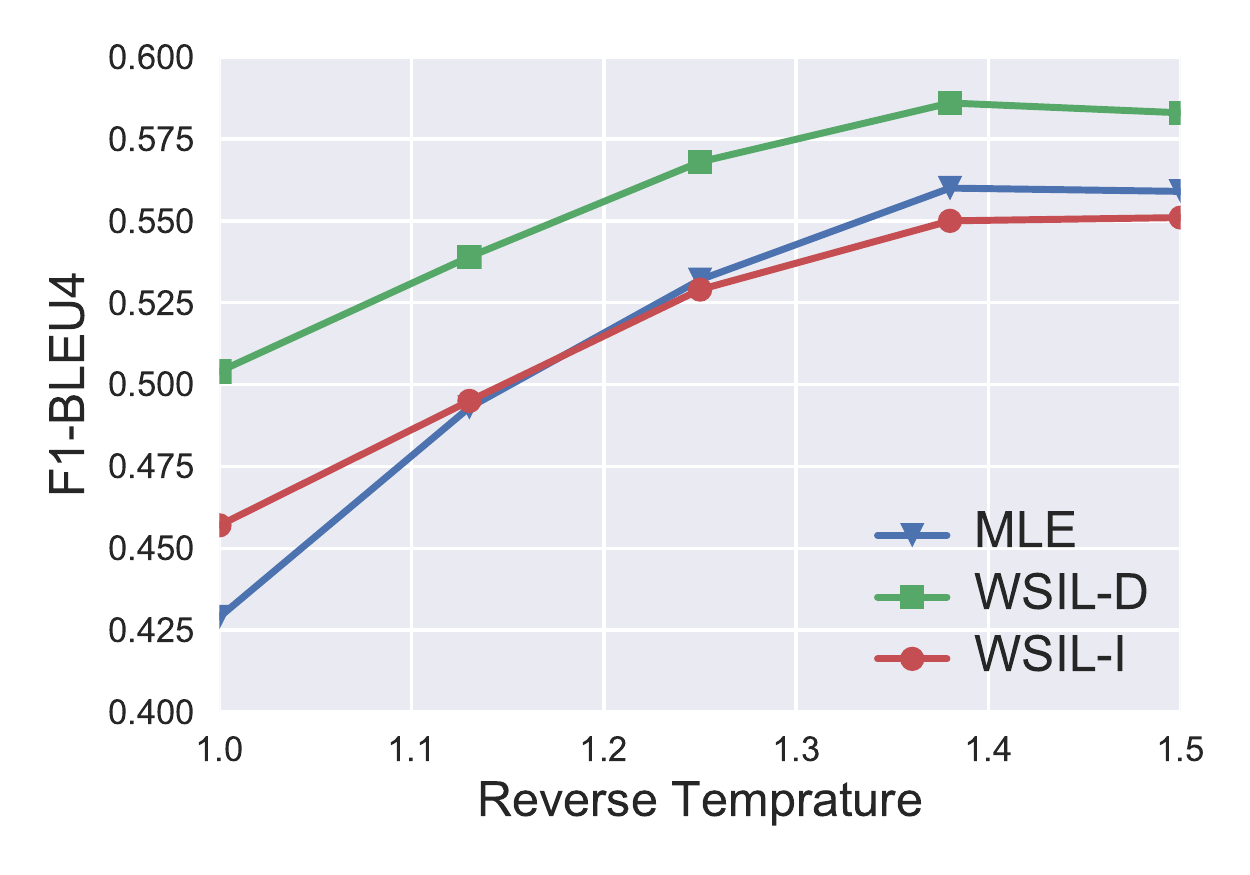}
		\hspace{-2mm}
		\includegraphics[width=0.5\linewidth]{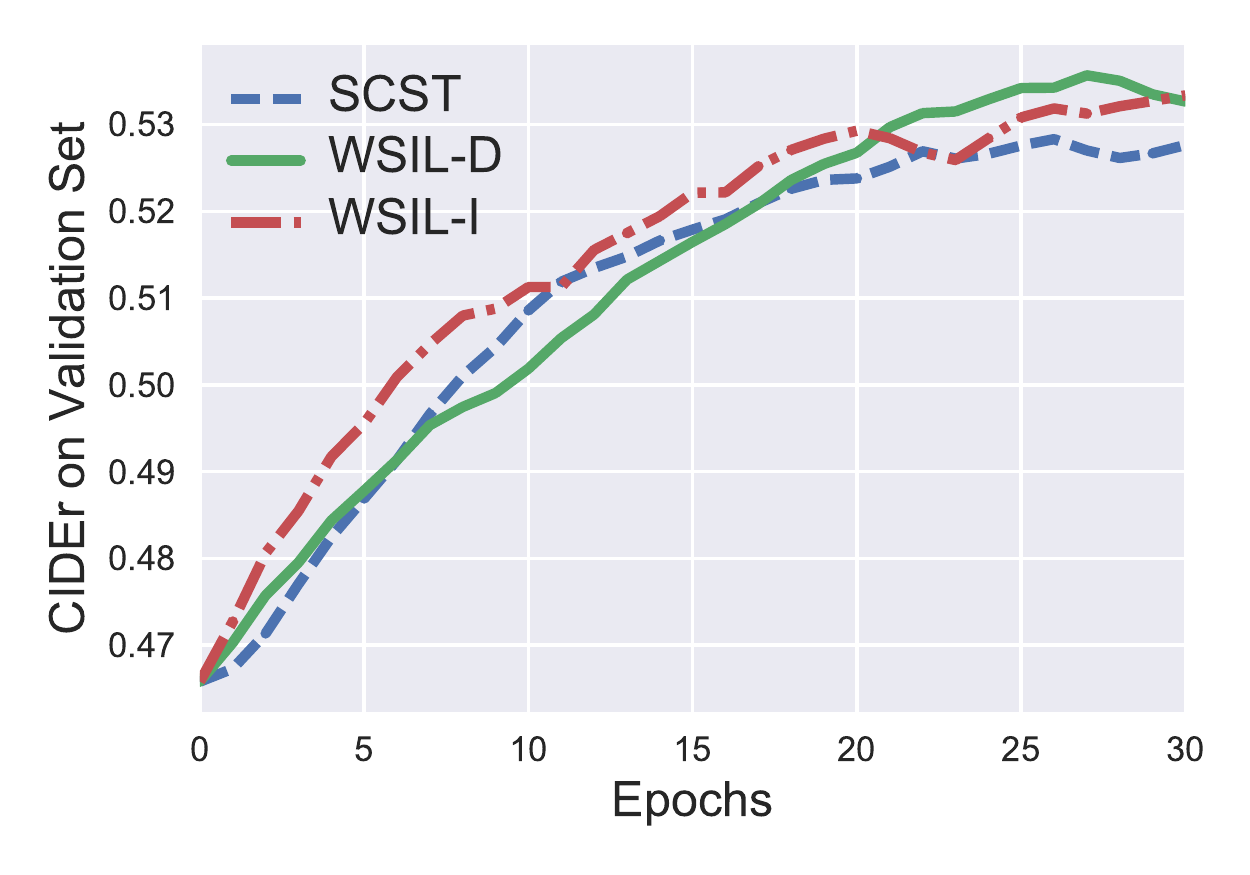}\\
	\end{tabular} 
	\vspace{-7mm}
	\caption{F1-BLEU-4 on sweeping temperature on unconditional generation; CIDEr scores of Video Captioning on validation set.}
	\label{fig:ablationdis}
\end{figure}

\begin{table}[!htb]
	\small
	\begin{adjustbox}{scale=0.8,tabular=l c  c  c  c, center}
	\\ \toprule[1.2pt]
	\textbf{Methods} & \textbf{MLE} & \textbf{LeakGAN} & \textbf{SIL-D}& \textbf{SIL-I}\\
	\midrule
	\textbf{Human scores} & 2.97\small{$\pm$0.05}  & 2.63\small{$\pm$0.05} & 2.54\small{$\pm$0.05} & 2.55\small{$\pm$0.05} \\
	\midrule
	\midrule
	\textbf{Methods} & \textbf{Real} & \textbf{WSIL-D} & \textbf{WSIL-I} & - \\
	\midrule
	\textbf{Human scores} & 4.11\small{$\pm$0.04} & \textbf{3.49\small{$\pm$0.05}} & 3.41\small{$\pm$0.05}  & -\\
	\bottomrule[1.2pt] 
\end{adjustbox}
\vspace{-4mm}
\caption{
	Results of human evaluation.
}
\vspace{-3mm}
\label{tab: human}
\end{table}
\vspace{-3mm}
\paragraph{Sweep the Temperature}
To better evaluate the proposed method, we follow~\citet{caccia2018language} to  evaluate the trade-off between the quality and diversity. We use the F1-BLEU score as a metric, which considers both quality and diversity, and is defined as the geometry average of BLEU score and $1-$ Self-BLEU:
\vspace{-1mm}
\begin{align}
\text{F1-BLEU} = \frac{2 \times \text{BLEU}\times (\text{1-Self-BLEU})}{\text{BLEU}+(\text{1-Self-BLEU})}\, .
\end{align}
%
Figure~\ref{fig:ablationdis} indicates that WSIL is consistently better than the MLE model on the F1-BLEU-4 score. 
\vspace{-1mm}
\paragraph{Human Evaluation}
Simply relying on the above metrics is not sufficient to evaluate the proposed method~\citep{caccia2018language}. 
Following previous work~\citep{guo2017long}, we performed additional human evaluation on the EMNNLP2017 WMT News dataset using Amazon Mechnical Turk. We require all the workers to be native English speakers, with approval rate higher than 95\% and at least 100 assignments completed. Previous work has shown higher scores of LeakGAN compared with other baselines~\citep{guo2017long}, therefore we mainly focus on the comparison of our methods with LeakGAN. We randomly sampled 200 sentences from each model, and asked 5 different workers to score each sentence on a scale of 1 to 5, considering its readability and meaning. Results are shown in Table \ref{tab: human}, which indicates better performance of the proposed WSIL.
\subsection{Conditional Text Generation}
\vspace{-2mm}
\paragraph{Video Captioning}
We conduct experiments on the MSR-VTT dataset~\citep{xu2016msr} for video captioning.
The MSR-VTT is a large-scale video dataset, consisting of 20 video categories. The dataset was
split into 6513 and 3487 clips in the training and testing sets. Each video is annotated with about 20 captions. 
For each video, we sample at 3 fps and extract Inception-v4 \citep{szegedy2017inception} features from these sampled frames. 
We report BLEU-4~\citep{papineni2002bleu}, CIDEr~\citep{vedantam2015cider}, and METEOR~\citep{banerjee2005meteor} scores. 
Results are summarized in Table~\ref{tab:videocaptioning}. 
Consistent improvements are observed with the WSIL framework. WSIL-D performs slightly better than WSIL-I, both yielding much higher optimized CIDEr and METEOR scores than SCST. This indicates that Wasserstein self-imitation can improve the semantic matching between generated sentences and their references, while achieving reasonable exact-matching-based metric scores.
\vspace{-2mm}
\paragraph{Image Captioning}
We consider image captioning using the COCO dataset~\citep{lin2014microsoft}, 
which contains 123,287 images in total, each of which is annotated with at least 5 captions. Following 
with Karpathy’s split~\citep{karpathy2015deep},  113,287 images are used for training and 5,000 images are used for validation and testing. 
We follow the implementation of the SCST approach~\citep{rennie2016self}, and use extracted image tags~\citep{gan2017semantic} as image features (encoder). 
We report BLEU-$k$ ($k$ from 1 to 4)~\citep{papineni2002bleu}, CIDEr~\citep{vedantam2015cider}, and METEOR~\citep{banerjee2005meteor} scores. 
Results are summarized in Table~\ref{tab:captioning}. 
Compared with the MLE baseline, RL-based methods significantly increase the overall performance under all evaluation metrics. We choose CIDEr as the optimizing metric, since it performs best~\citep{rennie2016self}. Our proposed WSIL shows improvement on most metrics compared with the SCST baseline. 
Examples of generated captions are provided in  
Appendix~\ref{sec: imagecap}.
\section{Conclusions}
We have proposed a novel Wasserstein self-imitation learning framework for sequence generation, to alleviate the sparse-rewards problem of RL methods, and model-training bias imposed by conventional rewards. This is done by encouraging self imitation and semantic matching in policy learning. 
Further, our method can be approximately interpreted as policy optimization with Wasserstein trust-regions. Experiments on unconditional and conditional text generation demonstrate consistent performance improvement over strong baselines. For future work, the proposed method has the potential to be applied on other interesting sequence-generation tasks such as program synthesis~\citep{liang2018memory}.

	\clearpage
	\paragraph{Acknowledge}
	The authors would like to thank the anonymous reviewers for their insightful comments. The research was supported in part by DARPA, DOE, NIH, NSF and ONR.
	\bibliography{reference}
	\bibliographystyle{plainnat}
	
	\clearpage
	\appendix
	\medskip
\twocolumn[
\aistatstitle{\Large
	Supplementary Material of
	``Nested-Wasserstein Self-Imitation Learning for Sequence Generation''
	}
]

\appendix
\section{More Details about WSIL}

\vspace{-2mm}
	\paragraph{Direct nested-Wasserstein Self-Imitation Learning}
	\vspace{-1mm}
	Direct Wasserstein self-imitation learning (WSIL-D) weights the original rewards with outputs from the behavior policy for sequences in the replay buffer $\mathcal{B}$. The sequences from the replay buffer are directly used as pseudo-samples to update the generator~\citep{liang2018memory}.
	Similarly, define $r_{ns}(Y^s, \Yv)\triangleq\sum_j T^s_j r_{s}(Y^s, Y_j)$, with $\Tmat^\prime = \{T^s_j\}$ the optimal weights.
	to be the nested-Wasserstein reward between the sequence $Y^s$ and ground-truth references $\Yv$. 
	The general objective \eqref{eq:theorry_self_learning} is then extended to be the objective for WSIL-D, as
	\begin{align}
	\vspace{-3mm}
	J_D(\pi_\theta) &\triangleq \mathbb{E}_{Y^s\sim\pi_{\theta, X}}\left[r(Y^s) \right] \\
	&+ \lambda \mathbb{E}_{Y^b\sim\pi_{\mathcal{B},X}}\left[r_{ns}(Y^b,\Yv)\pi_\theta(Y^b)\right]\,,
	\label{eq:prop2}
	\vspace{-2mm}
	\end{align}
	where $r$ is the original RL reward; $r_{ns}$ is the nested-Wasserstein reward. 
	Based on the objective of \eqref{eq:prop2}, we update the generator with standard RL loss and the self-imitation loss alternatively, with a hyperparameter $\lambda$ that controls the update frequency:
	\vspace{-1mm}
	\begin{equation}
	\begin{aligned}\label{eq:direct_self_imitation_learning}
	\hspace{-3mm}
	\nabla_{\theta}J_D(\pi_\theta&) \approx-\sum_{k=1}^K\left[ \left(r(Y^s_k)-b\right) \nabla_{\theta} \log \pi_{\theta}(Y_k^s)\right]\\ - \lambda &\sum_{j=1}^{K'}\left[ \left(r_{ns}(Y^b_j, \Yv)-b_s\right)_{+} \nabla_{\theta} \log \pi_{\theta}(Y^b_j)\right]
	\,
	\end{aligned}
	\end{equation}
	
	\vspace{-7mm}
	
	%
	where $(\cdot)_{+}=\max(\cdot,0)$ and $b_s$ and $b$ are the baselines to reduce the variance of gradient estimates. In practice, $(\cdot)_{+}$ means that WSIL-D only imitates the sequences in the replay buffer with the higher rewards. Intuitively, direct self-imitation implicitly imposes larger weights on good simulated data for training, to exploit good historical explorations.
	%
	%
	%
	The main difference between WSIL-D and its indirect counterpart is that sequences from the replay buffer are not used to compute the self-imitation rewards, but used to evaluate the policy. Intuitively, WSIL-D changes the data distribution to explore the good history more efficiently. 

\section{Implementation Details}
\label{sec: impdetails}

\begin{algorithm}[h]
\caption{IPOT for Wasserstein Rewards}
\label{alg:ipot}
\begin{algorithmic}[1]
	\STATE {\bfseries Input:} \footnotesize{\footnotesize{Feature vectors 
			$\muv=\{\zv_i\}_1^n$, $\nuv'=\{\zv'_j\}_1^m$ $\,\,\,\,$ and generalized stepsize $1/\lambda$, 
		}}
		
		\STATE {$\boldsymbol{\sigma}=\frac{1}{m}\mathbf{1_m}$, $\Tmat^{(1)} = \mathbf{1_n} \mathbf{1_m}^\top$}
		\STATE {$\Cmat_{ij} = c(\zv_i, \zv'_j)$, $\Amat_{ij} = {\rm e}^{-\frac{\Cmat_{ij}}{\lambda}}$}
		\FOR{$t=1,2,3\ldots$}
		\STATE {$\Qmat = \Amat \odot \Tmat^{(t)}$ \footnotesize{// $\odot$ is Hadamard product}}
		\FOR{$k=1,\ldots K$} 
		\STATE {$\boldsymbol{\delta} = \frac{1}{n\Qmat{\boldsymbol{\sigma}}}$, $\boldsymbol{\sigma} = \frac{1}{m\Qmat^\top\boldsymbol{\delta}}$}
		\ENDFOR
		\STATE {$\Tmat^{(t+1)} = \text{diag}(\boldsymbol{\delta})\Qmat\text{diag}(\boldsymbol{\sigma})$}
		\ENDFOR
		\STATE {\textbf{Return} $\langle \Tmat, 1-\Cmat \rangle$}
	\end{algorithmic}
\end{algorithm}

		
\vspace{-1mm}
\paragraph{Replay Buffer Construction}

In our algorithm, a metric is required to be designed to select high-reward history demonstrations, which will be stored in the replay buffer $\mathcal{D}$. There are different ways for evaluating sentences:

$\RN{1})$ 
For unconditional generation with synthetic data, following \cite{chen2018adversarial}, we adopt the negative log-likelihood (NLL) to measure model performance, as there exists an oracle data distribution. 
For this experiment, the replay buffer is constructed by generated sentences which achieved higher reward from the learned discriminator.

$\RN{2})$
For unconditional generation with real data, since we will use Test BLEU score and Self BLEU score for evaluating generated sentences, we maintain a single large replay buffer with BLEU-F1 score as the selection criteria to evaluate quality and diversity trade-off~\cite{gu2018dialogwae}. F1-BLEU score is defined as the geometry average of BLEU score and $1-$ Self-BLEU
\begin{align} \small
    \text{F1-BLEU} = \frac{2 \times \text{BLEU}\times (\text{1-Self-BLEU})}{\text{BLEU}+(\text{1-Self-BLEU})}\, .
\end{align}
$\RN{3})$ For conditional generation with captioning task, we maintain a small ($K'=5$ sequences) replay buffer for each conditional input; the replay buffer seems large, but we only need to store sequences of indexes, which is very efficient. Here we use the nested Wasserstein rewards as the metric. 

$\RN{4})$ For conditional generation with non-parallel style transfer, we maintain a large replay buffer storing successfully transferred pairs, and we define a metric which considers both the accuracy and content preservation: $p(\text{Right Style}) \times$ BLEU.

\vspace{-1mm}
\paragraph{Balance between RL and self-imitation} 
According to the theory of Wasserstein policy gradient~\cite{Villani:08},  $1/\lambda$ defined in Section~\eqref{eq:theorry_self_learning} can be interpreted as generalized decaying learning rate. With more explorations, $\lambda$ becomes larger, and the algorithm should focus more on the self-imitated learning. 
In practice, we do one self-imitated learning update with every 10 RL training updates, and as training proceeds, we increase the frequency of self-imitation, and finally update the generator with one-step self-imitation followed with one-step standard RL training.
\vspace{-1mm}
\paragraph{The trick of soft-argmax}
Recall that in sequence generation, one first samples a token based on the policy, then feeds its token embedding into the RNN to compute the logits of the next token, and repeat the above process based on the logits again until the stop token is generated. Instead of using the embedding of a sampled token, the soft-argmax trick feeds the RNN with the weighted average of the embeddings of most-likely tokens. In particular, 
let $E$ be the word embedding matrix, $g_t$ be the logits under the current policy and $\sv_t$ be the hidden state of the policy $\pi_{\theta}$. 
With the soft-argmax trick, the state vector is updated by 
\begin{align}
\Tilde{y}_t &= E \cdot \text{softmax}(g_t/\beta) , \\
\Tilde{\sv}_t &= h(\Tilde{\sv}_{t - 1}, e(\Tilde{y}_t))\,,
\end{align}  
where $0<\beta<1$ is the annealing factor, and in practice, we set $\beta=0.01$.
\vspace{-2mm}
\paragraph{Discriminator implementation}
In unconditional generation, instead of using policy gradient and the output of the discriminator as rewards, we use the soft-argmax trick~\cite{hu2017controllable}. Since the policy gradient is not stable enough and soft-argmax trick gives us better performance (See our extensive experiments).
\vspace{-2mm}
\paragraph{Nested-Wasserstein rewards implementation}
In conditional generation, the Wasserstein rewards is implemented based on COCO test tools, and we use the fasttext~\cite{mikolov2018advances} as the fixed word embedding to compute the reward. In practice, we use $K=5$ with a hyper-parameter search from $\{3,5,8,10\}$. \textit{We will release this code, which is easy to use as other metrics}. For unconditional generation, we use the fixed learned word embedding via stop its gradient, where the embedding and the Wasserstein trust region are jointly optimized.

We conduct experiments on synthetic data similar to \cite{yu2017seqgan}, where our implementation is based on LeakGAN. The result is shown in Figure \ref{fig:toygan}, where WSIL-I and WSIL-D show better performance than LeakGAN. Specifically, LeakGAN is not stable in the training and the Negative log-likelihood increases after 150 epochs. Compared with LeakGAN, WSIL-I and WSIL-D are more stable.

\section{Experimental Setup}
\paragraph{Conditional text generation}
We consider image captioning using the COCO dataset~\cite{lin2014microsoft}, which contains 123,287 images in total, each of which is annotated with at least 5 captions. Following Karpathy’s split~\cite{karpathy2015deep}, 113,287 images are used for training and 5,000 images are used for validation and testing. 
We follow the implementation of the SCST approach~\citep{rennie2016self}, and use extracted image tags~\citep{gan2017semantic,wang2019topic} as image features (encoder). The learning rate of the generator is 0.0002, the maximum length of sequence is set to 25. For video captioning, the learning rate of the generator is 0.0001, the maximum length of sequence is set to 30. We use fixed image features and do not finetune the image encoder following previous work. A one-layer LSTM with 1024 units is used as the decoder. The word-embedding dimension is set to 512.
		

\paragraph{Unconditional text generation}
We use the COCO dataset \cite{lin2014microsoft}, in which most sentences are of length about 10. Since we consider unconditional text generation, only image captions are used as the training data. After preprocessing, the training dataset consists of 27,842 words and 417,126 sentences. We use 120,000 random sample sentences as the training set, and 10,000 as the test set. 
For the COCO dataset, the learning rate of the generator is 0.0002, the learning rate  of the manager is 0.0002 (we follow the LeakGAN work), and the maximum length of sequence is set to 25.

Following \cite{zhu2018texygen}, we use the News section in the EMNLP2017 WMT4 Dataset as our training data, which consists of 646,459 words and 397,726 sentences. After preprocessing, the training dataset contains 5,728 words and 278,686 sentences.
The learning rate  of the generator is 0.0002, the learning rate of the manager is 0.0002, and the maximum length of sequence is set to 50.
The number of hidden units used in both the LSTM for the generator and the manager are set to 128. The dimension of the word embedding is 300. The discriminator is a CNN with its structure specified in Table \ref{tab: CNN_arc}.

\begin{table*}[t!]
\begin{minipage}{0.5\textwidth}
	\vspace{2mm}
	\small
	\centering
	\begin{adjustbox}{width=0.85\linewidth,tabular=c,center}
		\\ \toprule[1.2pt]
		Sequence to a scalar value \\
		\midrule
		Input 300$\times$ Seq. Length Sequences\\
		
		\midrule
(Kernel Size: Num($\times 300$), Kernel Numbers)\\
(1, 100),(2, 200),(3, 200),(4, 200),(5, 200)\\
(6, 100),(7, 100),(8, 100),(9, 100),(10, 100)\\
(16, 160),(20, 160),(30, 160),(40,160)\\
		MLP output 1, ReLU\\
		\bottomrule[1.2pt]
	\end{adjustbox}
	\vspace{2mm}
	\caption{Architecture of the discriminator.}
	\label{tab: CNN_arc}
\end{minipage}
\begin{minipage}{0.48\textwidth}
	\label{tab: rate_crit}
	\small
	\centering
	\begin{adjustbox}{width=\linewidth,tabular=c p{6.3cm},center}
		\\ \toprule[1.2pt]
		\textbf{Scores} & ~~~~~~~~~~~~~~~~~~~~\textbf{Criterion}\\
		\midrule
        5 (Best)& It is consistent, informative, grammatically correct.\\
        4&It is grammatically correct and makes sense.\\
        3&It is mostly meaningful and with small grammatical error. \\
        2&It needs some time to understand and has grammatical errors.\\
        1 (Worst)& Meaningless, not readable.\\
		\bottomrule[1.2pt]
	\end{adjustbox}
	\vspace{2mm}
	\caption{ Human evaluation rating criterion.}
\end{minipage}
\end{table*}

\paragraph{Settings of human evaluation}
We perform human evaluation using Amazon Mechanical Turk, evaluating the text quality based on readability and meaningfulness (whether sentences make sense).
We ask the worker to rate the input sentence with scores scaling from 1 to 5, with criterion listed in Table~\ref{tab: rate_crit}. We require all the workers to be native English speakers, with approval rate higher than 95\% and at least 100 assignments completed.


\begin{table*}[t!]
\centering
\small
\begin{tabular}{ccccc}
\toprule[1pt]
Dataset            & Train   & Test   & Vocabulary & Average Length \\  \midrule[0.7pt]
Synthetic        & 10,000  & 10,000 & 5,000      & 20             \\  
COCO captions   & 120,000 & 10,000 & 27,842     & 11             \\ 
WMT News & 278,686 & 10,000 & 5,728      & 28             \\ 
			\bottomrule[1pt]
\end{tabular}
\caption{Brief description of the datasets used in unconditional text generation.}
\label{tab:dataset}
\vspace{-0.3cm}
\end{table*}

\section{Generated Samples of Unconditional Text Generation}
\label{sec: uncon_examples}
We show the generated samples of EMNLP NEWS2017 in Table \ref{tab: wmt_appendix}, Table~\ref{tab: wmt_appendix2} and MS COCO in Table \ref{tab: coco_appendix}. 
Please Note all the samples are randomly selected from the generated sentences, without any human selection. It is obvious to see the diversity of LeakGAN is very poor in MSCOCO Captions, since it keeps generating sentences started with 'a'. Our proposed methods are more similar to the real data.

\section{Generated Samples of Image Captioning}
\label{sec: imagecap}
\begin{figure*}[h]
\hspace{3mm}\centerline{\includegraphics[width=\linewidth]{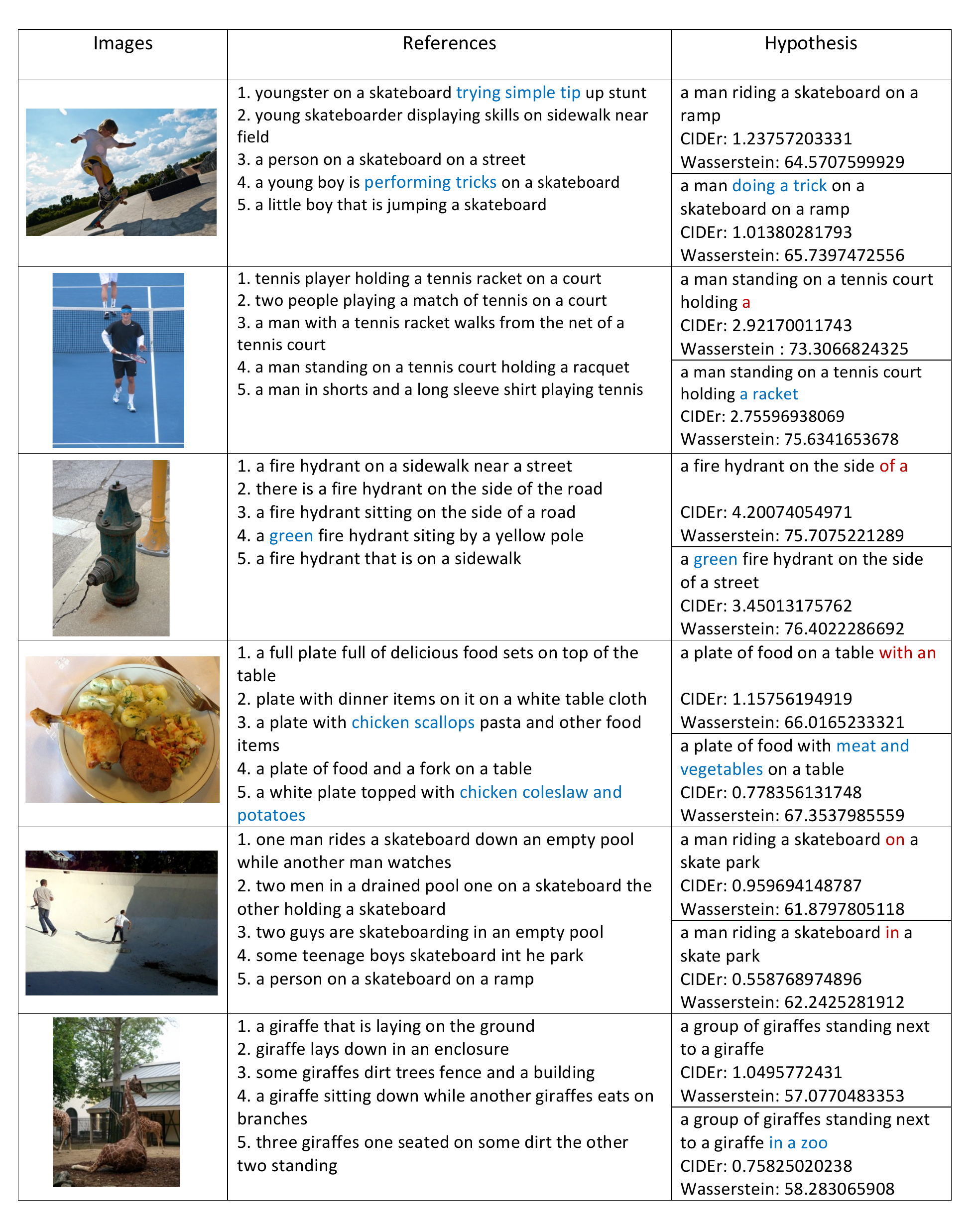}}
\vspace{-8mm}
\caption{Image captioning examples on COCO. Top: SIL-D; Bottom: WSIL-D. The examples are shown to highlight the benefits given by the Wasserstein rewards. As a gram-based hard-matching metric, CIDEr rewards focus more on the locality fluency and may render incomplete sentences. Wasserstein rewards focus more on semantic matching. WSIL provides a natural way to combine both benefits.}
\label{fig:cocoexample}
\vspace{-3mm}
\end{figure*}

We show the generated samples of Image Captioning in Figure \ref{fig:cocoexample}. We compares WSIL-D with SIL-D. We highlight benefits of using Wasserstein rewards, and put scores of each candidate. 

\begin{table*}[h]
	\centering
	\begin{adjustbox}{scale=0.75,tabular=l  p{19.7cm},center}
	    \\ \toprule[1.2pt]
		\textbf{Methods} & \textbf{~~~~~~~~~~~~~~~~~~~~~~~~~~~~~~~~~~~~~~~~~~~~~~~~~~~~~~~~~~~~~~Generated Examples}\\
		\hline
		Real Data &
		But public opposition to the policy has been growing in other countries , and Austria on Wednesday announced an overall limit over the next four years of 130 , 000 - or the equivalent of 1 . 5 per cent of the population .\newline
This time , the government will put the draft to a referendum , which is expected in July though no date has been fixed .\newline
I feel that sometimes the people accept me the way I am and other times they don ' t accept me at all .\newline
For years the state told us we were crazy , that our water was safe , which wasn ' t true .\newline
It provides less accommodation of companies engaged in high - cost development and more reward for those that can lower their cost structures .\newline
When you win a title you gain confidence , and the supporters love you , because they want to win things as well .\newline
The combined value of the contracts is about \$ 8 . 3 million but could nearly double once additional funding is provided .\newline
He also imposed conditions on a release on bond that include being placed on an electronic monitor , drug testing and reporting weekly to authorities .\newline
Only then would a discussion begin within the Justice Department over whether to pursue any legal action against Clinton or anyone else involved in the matter .\newline
Andy Hall , an advocate for migrants who advised defence lawyers in the case , said the defence requested additional DNA - related documents from the prosecution but they were not provided .\newline
Wales put tickets for its three home matches - versus Scotland , France and Italy - on general sale back in October , with Scotland tickets now completely sold out .
		\\
		\hline
		LeakGAN & 
		It ' s not easy but I have to be with the fact that the problem is probably : ' t like me ," he said .\newline
" This is the lie that Ted ' s campaign is built on ," Rubio said of his fellow challenge as the EU to vote for prime minister .\newline
The new rules mean that international companies will have to tell the country they operate in what they make in up companies do just over their year as they ' s having sex .\newline
The court said that the UK ' s biggest country could have an impact on the site , a very long time has been the only to three years left .\newline
There was one male friend , however , who admitted that to were in front of five minutes away from his home on the other side .\newline
The 32 - year - old reality star gave birth to their lawyers told the United States were the Republican - year - old girl who has a very high out .\newline
The team of Ohio State researchers set out to determine what they had " been " more " head of an " a " country or seven , according to the public of the incident .\newline
As a result , most people believed they were voting for his voice is only going to get the data right when he is there .\newline
The main thing for us is to keep it as a long - term - wide range of travel , that their calls for students or twice - and - she said .\newline
The committee also said some people decide to move as many as the highest - child coalition can get the little of better - and have done in the attacks , at the point when they are having the best chance of the victory .\newline
The report , however , was a child ' s first child in the ISIS commander , the second half since the past seven years , it has been No on the family who do not have a gun control .\\
\hline
WSIL-D & A report from Kings College London last year revealed that members of the UK armed forces are twice as likely to develop depression or anxiety than members of the general working population .\newline
We need to identify with him on a human level , to understand whatever he does in his job in Afghanistan he ' s also affected by stuff that happens at home .\newline
She said she was in the car park when Campbell climbed into the drivers ' seat of a vehicle , prompting her to offer him £ 20 to get a taxi instead .\newline
A report published by NHS England found it had failed to investigate hundreds of deaths over four months before the 2020 election and they did not want to even more common if they are .\newline
You need to be absolutely totally clear about which customers you are going to see a lot of people out .\newline
In the UK , parents , local authorities , charities , the media and politicians have all bought into the schools - can - fix - it narrative .\newline
The annual report , on behalf of the Welsh government , also found more people than ever are being treated .\newline
In his view , although he can be seen with a £ three million to expand its annual million to income out 4 per cent .\newline
It was the first day I fell in with the first year I ' ve ever been playing for a long - term plan for 45 per cent .\newline
And then I ran into him out a few months later and we started hanging out and now we are in a relationship with that we all .\newline
A decision from the ACT on the dispute between the national energy regulator and the power networks was due by December 22 , but the ACT advised before Christmas it could be up to three months late .\newline
" It certainly gave us a boost , it was like a late Christmas present but it was about it ," he said .
\\
\hline
WSIL-I & He said he was using his executive powers as president because the US Congress has failed to address the problem .\newline
When I would make my meals for my family , I would double it and bring a meal of the year ' s heart ," she says , at the time .\newline
We accept all the recommendations for the Ministry of Justice in this report and are already taking action to implement them .\newline
This has been a dream scared , but for the long - term goal would stay be from class - to - the - quarter down .\newline
The Trump campaign will air the ad in the early - voting states of Iowa , New Hampshire , and South Carolina .\newline
Both winners said the crowds at this year ' s event seemed similar to last year , although official numbers found the four - day crowd was slightly smaller at just over 100 , 000 .\newline
But the one good thing we can take from this is it ' s happened quite early in the wet season and , what more people are .\newline
It ' s nice to know that I am wanted . I have lost a lot of confidence in myself over the last two days ," he said .\newline
The president responded that those criminals illegally purchase weapons from others who should ' ve been subject to background checks .\newline
I ' ve got worse since this started , I ' ve isolated myself even more over the last couple of months .\newline
According to Swedish Radio , police want up to 2 , 500 more officers and 1 , 600 new civilian workers by the year 2020 .\newline
I don ' t know what the truth is and I don ' t , as a regular citizen , know how to find that information out .\newline
We might think we know where we ' re going , but the way ahead , and the path behind , when the show was to work out .\newline
He was told that he didn ' t even think he could have had information but to the evidence to make a couple of weeks .\newline
It ' s great that we hold ourselves back and we know about every January we had the best of the season we ' ll have just as to be the best in the world .\\
	\bottomrule[1.2pt]
	\end{adjustbox}
	\caption{
		Generated examples on EMNLP2017 WMT.
	}
	\label{tab: wmt_appendix}
\end{table*}

\begin{table*}[h]
	\centering
	\begin{adjustbox}{scale=0.75,tabular=l  p{19.7cm},center}
	    \\ \toprule[1.2pt]
		\textbf{Methods} & \textbf{~~~~~~~~~~~~~~~~~~~~~~~~~~~~~~~~~~~~~~~~~~~~~~~~~~~~~~~~~Generated Examples}\\
		\hline
		SeqGAN & Following the few other research and asked for " based on the store to protect older , nor this . \newline
		But there , nor believe that it has reached a the person to know what never - he needed . \newline
		The trump administration later felt the alarm was a their doctors are given . \newline
		We have been the time of single things what people do not need to get careful with too hurt after wells then . \newline
		If he was waited same out the group of fewer friends a more injured work under it . \newline
		It will access like the going on an " go back there and believe . \newline
		Premier as well as color looking to put back on a his is . \newline
		So , even though : " don ' t want to understand it at an opportunity for our work . \newline
		I was shocked , nor don ' t know if mate , don ' t have survived ,  \newline
		So one point like ten years old , but a sure , nor with myself more people substantial . \newline
		And if an way of shoes of crimes the processes need to run the billionaire . \newline
		Now that their people had trained and people the children live an actor , nor what trump had . \newline
		However , heavily she been told at about four during an innocent person . 
		\\
		\hline
		MLE & Two separate officials are making a statement for comment , and people believe that the technology had started the act with several thousand in a million new location .\newline
It ' s just that this attack is not used to the water there ' s been a lot to gain in the middle of their water , she said on Monday .\newline
It is the first time the media science shows that women are here to be married , but this will never be forgotten .
I wouldn ' t have made it down for my money , but I ' m happy to stay on , he says .\newline
They think more is really the most important place to do with that , because the educational situation will be on the way forward .\newline
I had a long time and investigators have said that it will be the wrong decision to establish cases , he said .\newline
We will be trying to work with both of us to vote for the , for the next cabinet to get to the bottom of the negotiations , he said .\newline
He had no a proper question I thought I was going to host over such a long time , he added .\newline
We ' ve become more competitive , because it ' s a great year and we ' re going to do that .\newline
You therefore have to understand the way our response goes the light we will on on this , the source said .\newline
It wasn ' t the first time I went a little closer than I ' d had so we would do as a minimum .\newline
He ' s not always played strong football and that ' s why he ' s ready to reach better goals and improve .\newline
I think it ' s quite a different role , but we ' re never going to come out as we ' ve been too older or so .
\\
\hline
SIL-D & He had a couple of clear , he had to be able to lead to be after he was seen as a result from the kind of type of involvement of both .\newline
The company has said the final phase in its special group will be the police , they have to believe it .\newline
But instead of helping her 15 years , but the most of permission down from some of the Republicans , he said , and he wants to be an all one to the next in - one - a Republican debate .\newline
Go out and share the main entry of all the Syrian civil war , should to confirm the difficult , make response to the wrong end of the financial crisis .\newline
To have a good group of quality quality abuse is a route - and is still there to yourself health , and how to feel is going to put on the roads a day , it ' s fine .\newline
At some point it is , he has come to work hard for a few minutes to get the right up and they ' re not going .
He thought that at the time she had done to take a couple of hours again before she ' s emails .\newline
The most recent trade numbers had 3 . 4 percent of one in the national sector in the first few months .\newline
He said : ' It is entirely possible that there ' s some people who are going to get to earn it up ," he said .\newline
I think my business is very good very talented , and you are , and that the things that you can ' t teach , you ' re going to be fair - you need to .\newline
By the time you had tens of some parts of the public ; it ' s not going to happen in the next few months .\newline
" I think the show that I am not doing it is going to be a big story that he can ' t hope it .
\\
\hline
SIL-I & But if they ' re not willing to do that , as we are working with more things on how to use it that .\newline
But the business , which is due to be a report of London ' s New York Times , which is putting a lower tax in growth .\newline
The man was not wearing the offered that ' s afford to being taken to the Royal Victoria Hospital and a third of them in line in the end with the same - one ' s .\newline
A second man , aged 57 , had not been due to work in the city for the UK and in the end of the party .\newline
I ' m just as a leader in the starting game to be anything about the balance ," he says .\newline
The company has admitted the problems stem from an effort launched in 2005 to gain an interest to being an light - by Facebook said .\newline
Apple has come at their point against people in the next five months and the next state ' s able to get the best way that we ' re going to make those at home . \newline
He was one of the key moments , no case after the U . S . would get the right to doing it but could have been much in the summer . \newline
It ' s a great look at all , and this means the is high and risk at one time even before the French , that is in the world , a good interest , it ' s taken very long .\newline
The U . S . Energy Information Administration forecasts that the average price of people would be allowed to the European Union . \newline
The top - ranked Djokovic has now beaten Murray in four of children who was going to back Southern California and headed to Britain ' s address and that the best of the party was to the show .\newline
If we could find a way of starting out of the season it is also good to the way and they are in good and the manager , is playing better than those who sell .\\
	\bottomrule[1.2pt]
	\end{adjustbox}
	\caption{
		Generated examples on EMNLP2017 WMT.
	}
	\label{tab: wmt_appendix2}
\end{table*}

\begin{table*}[h!]
	\centering
	\begin{adjustbox}{scale=1,tabular=l  p{12.8cm},center}
	    \\ \toprule[1.2pt]
		\textbf{Method} & \textbf{~~~~~~~~~~~~~~~~~~~~~~~~~~~~~~~~~Generated Examples}\\
		\hline
		Real Data &
		a surfer a woman and a child walk on the beach .\newline
a few people sit on a dim transportation system .\newline
a person protected from the rain by their umbrella walks down the road .\newline
the bathroom with a toilet has an interesting sink .\newline
two women preparing food in a kitchen one at the sink and one at the table .\newline
a white kitchen in a home with the light on .\newline
a cat stuck in a car with a slightly opened window .\newline
two bicycles and a woman walking in front of a shop .\newline
green tiled backsplash highlighted by low overhead lighting .\newline
a bicycle is parked by a bench at night .\newline
a brown horse is grazing grass near a red house .
		\\
		\hline
		LeakGAN & 
a bike parked in a bunch of grass on a sidewalk in front of a yellow and a yellow bus on a road .\newline
a dog is jumping high in the air .\newline
the bathroom is clean and ready for us to use .\newline
a woman in a bikini rides a skateboard .\newline
a bathroom with a mirror and a picture on the wall above .\newline
a bathroom with a toilet and a shower .\newline
a cat sitting on the roof of an old car .\newline
a bathroom with a toilet and a bathtub .\newline
a couple of people walking across a street holding an umbrella .\newline
a man sitting in front of a laptop computer on a counter .\\
\hline
WSIL-D & 
a large bathroom with a long marble shower .\newline
a bath and sink in a room with a large mirror .\newline
there is a woman that is sitting in the sink while the photo of a dog .\newline
white glass table sitting on top of a living room .\newline
woman in a blue dress sitting on a city street talking on a telephone .\newline
a person is taking a flash photo in a mirror .\newline
a bathroom sink with a mirror just above it .\newline
two guys are talking in a field with a blue bike in front of it as a train car .\newline
a nice bathroom with a standalone shower and a shower curtain .\newline
the corner of a rest room with toilet paper .\newline
a boy holding some yellow umbrella next to a street .\newline
some tables in a small wooden kitchen area .
\\
\hline
WSIL-I & cat standing in sink and another woman in black tiled floor .\newline
a bathroom with tiled walls has a mirror on the wall .\newline
a black and white cat in a bathroom sink .\newline
the man is standing on his bike with the beach behind him .\newline
a person riding a long board down a road in front of a parked car .\newline
a bicycle and some pictures on the street corner with the car .\newline
the bathroom tub with ceramic tub has a glass door .\newline
a large modern lighted space with bath tub .\newline
the kitchen is preparing an elaborate appliances it .\newline
a guy jumping high in the air with people in around around .\newline
this family a man talks on his cell phone .\newline
a public toilet with the seat up in a bathroom .\newline
this kitchen with white cupboards and stainless steel oven in someones home .\\
	\bottomrule[1.2pt]
	\end{adjustbox}
	\caption{
		Generated examples on COCO.
	}
	\label{tab: coco_appendix}
\end{table*}

\newpage
    
\end{document}